\documentclass[letterpaper]{article} 
\usepackage{aaai2027}  
\usepackage[hyphens]{url}  
\usepackage{graphicx} 
\usepackage{natbib}  
\usepackage{caption} 
\usepackage{algorithm}
\nocopyright
\usepackage{newfloat}
\usepackage{listings}
\DeclareCaptionStyle{ruled}{labelfont=normalfont,labelsep=colon,strut=off} 
\floatstyle{ruled}
\newfloat{listing}{tb}{lst}{}
\floatname{listing}{Listing}
\usepackage{microtype}
\usepackage{multirow}
\usepackage{booktabs}
\usepackage{xcolor}
\usepackage{fvextra}
\usepackage{inconsolata}

\usepackage{graphicx}
\usepackage{amsmath}
\usepackage{amssymb}
\usepackage{algorithm}
\usepackage{algpseudocode}
\usepackage{fancyvrb}
\usepackage{booktabs}
\newcommand{\qwenicon}{\raisebox{-0.2\height}{\includegraphics[height=1.1em]{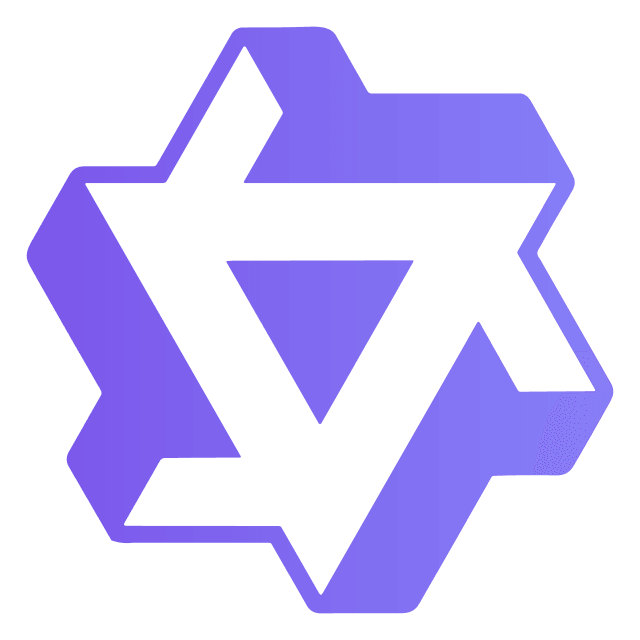}}}
\newcommand{\llamaicon}{\raisebox{-0.2\height}{\includegraphics[height=1.1em]{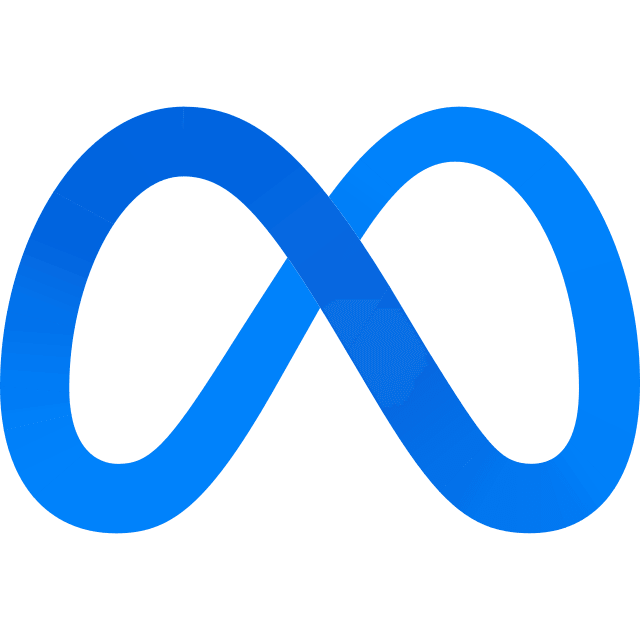}}}
\newcommand{\deepseekicon}{\raisebox{-0.2\height}{\includegraphics[height=1.1em]{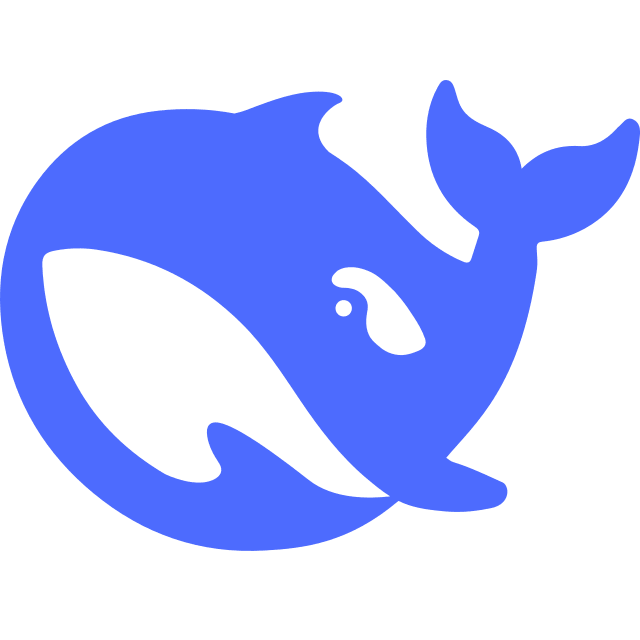}}}
\title{Rewarding Better Thinking for LLM Preference Alignment}
\author{
    Xubo Liu\textsuperscript{\rm 1}, Wenya Guo\textsuperscript{\rm 1}, Ruxue Yan\textsuperscript{\rm 1}, Xinying Qian\textsuperscript{\rm 1}, Ying Zhang\textsuperscript{\rm 1}
}
\affiliations{
    \textsuperscript{\rm 1}VCIP, DISSec, College of Computer Science, Nankai University, Tianjin, China \\
    \{liuxubo, guowenya, yanruxue, qianxinying\}@dbis.nankai.edu.cn, yingzhang@nankai.edu.cn

}

\begin{document}

\maketitle

\begin{abstract}
LLM preference alignment aims to optimize models toward human preferences across diverse user instructions. 
Reinforcement learning has become a major post-training approach for this goal, but existing proxy rewards are often outcome-level, mainly evaluating the final response while providing limited guidance for the reasoning trajectory. 
This can make credit assignment coarse when multiple responses receive similar final scores, leaving trajectory-level preferences under-specified. 
To address this limitation, we propose Thinking Checklist Reward (TCR), a process-oriented reward for RL-based preference alignment. 
TCR converts preference pairs into sample-specific thinking checklists and uses them to evaluate whether the generated reasoning trace addresses the preference-implied considerations. 
To reduce overlap with outcome-level supervision, TCR further introduces an exponential moving average (EMA) residual formulation to isolate a complementary thinking surplus beyond what is predictable from the outcome reward. 
Experiments on five models from three model families show that TCR consistently improves alignment performance across diverse benchmarks, with ablations further validating the importance of EMA-based residual formulation and sample-specific checklist supervision.
\end{abstract}


\begin{figure}[t]
    \centering
    \includegraphics[width=\linewidth]{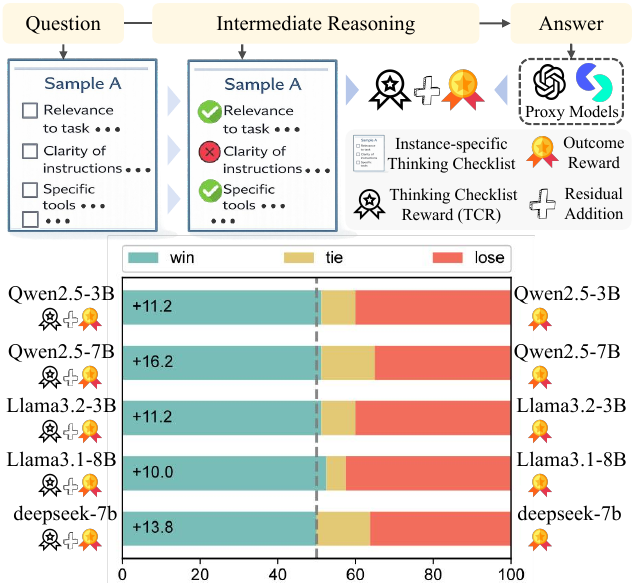}
    \caption{(Upper) For each input, a sample-specific thinking checklist is inferred from preference pairs, the generated reasoning trace is evaluated against the checklist, and the resulting process-level signal is combined with the outcome reward through a residual formulation. (Lower) On Vicuna Eval’s pairwise evaluation, training with TCR consistently improves alignment performance across five models.}
    \label{fig:intro}
\end{figure}

\section{Introduction}
Reinforcement learning (RL) has emerged as a major post-training paradigm for large language models (LLMs) \cite{yang2025qwen3,guo2025deepseek}. In domains with objective verifiers, such as mathematics and coding, reinforcement learning with verifiable rewards (RLVR) has become a central approach to training reasoning models, exemplified by GRPO \cite{shao2024deepseekmath}. Building on this success, recent works \cite{wang2025autorule,xu2025direct,zhang2025auditable,simonds2025rlsr,su2025crossing,shen2026rethinking,ou2026serl} have begun to extend reinforcement fine-tuning to open-ended alignment using proxy reward signals, often provided by reward models \cite{liu2024skywork} or LLM judges \cite{zheng2023judging}.

Despite these advances, most proxy rewards used in open-ended alignment remain predominantly outcome-level: they assess the final response while providing little direct signal about the generation trajectory. 
This can limit credit assignment in reasoning-oriented LLMs, especially when multiple responses receive similar final-response scores. 
In such cases, outcome-only rewards provide limited guidance on which trajectories better reflect the user's intent, constraints, and preference-implied considerations. 
As a result, useful reasoning patterns discovered during exploration may be under-credited, even when they could support more robust and well-aligned behavior.

Process Reward Models (PRMs) have been proposed to address a related limitation in mathematics and coding by providing supervision over intermediate reasoning steps \cite{chen2024step,chen2025reasongrm,ye2025uncertainty,he2025good,zou2025reasonflux,fan2025posterior}. 
Such supervision is effective in these domains where intermediate steps can often be locally evaluated according to correctness or their contribution to solving the problem. 
However, directly extending these PRMs to open-ended alignment is challenging \cite{zhang2025openprm,tian2026rit}. 
Unlike mathematics or coding, open-ended tasks often lack a unique solution path or locally verifiable intermediate steps. 
The value of a reasoning trace may instead lie in whether the overall trajectory captures the user's intent, relevant constraints, trade-offs, and contextual appropriateness, rather than whether each individual step is independently correct. 
Such trajectory-level considerations guide how the final response selects information, balances constraints, and satisfies preference-implied requirements.

To this end, we propose the Thinking Checklist Reward (TCR), a process-oriented reward component for open-ended alignment. 
As illustrated in Figure~\ref{fig:intro}, TCR first infers a sample-specific thinking checklist from preference pairs by extracting the key considerations that make the preferred response better. 
It then evaluates the generated reasoning trace against this checklist and converts the evaluation into a trajectory-level reward signal. 
In this way, TCR transforms preference information into explicit supervision for the reasoning process, helping distinguish trajectories that receive similar outcome-level scores but differ in their underlying reasoning quality.

A further issue is that reasoning quality and final-response quality are often positively correlated. 
As a result, directly adding a checklist-based reward may introduce reward redundancy by reinforcing information already captured by the outcome reward. 
To mitigate this reward overlap,
TCR adopts a residual formulation based on an exponential moving average, which estimates the predictable component of checklist quality from the outcome reward and retains only the remaining thinking surplus. 
This design makes TCR complementary to outcome-level supervision and allows it to be incorporated into standard RL post-training frameworks.
We evaluate TCR on five models from three model families and show consistent improvements across diverse benchmarks. 
Our contributions are summarized as follows:
\begin{itemize}
    \item We propose TCR, a preference-derived checklist reward that evaluates reasoning traces in a sample-specific and trajectory-level manner.
    \item We introduce a residual reward formulation that reduces overlap with outcome-level supervision and isolates a complementary thinking surplus for RL post-training.
    \item We empirically show that our sample-specific thinking checklist outperforms unified global checklists and generic trajectory scoring.
\end{itemize}

\section{Related Work}

\paragraph{RL Post-training for LLM Alignment.}
RL has become a central post-training paradigm for LLM alignment, from RLHF-style preference optimization to RLVR for reasoning tasks \cite{shao2024deepseekmath,guo2025deepseek,yu2025dapo}. 
Recent work extends RL to open-ended alignment with reward models or LLM judges as proxy rewards \cite{liu2024skywork,zheng2023judging,wang2025autorule,xu2025direct,zhang2025auditable,zhong2026p2s}. 
However, these signals mostly evaluate final responses, so they provide limited guidance when different reasoning trajectories lead to similarly rated answers. 
TCR instead derives trajectory-level supervision from preference pairs, encouraging desirable reasoning behaviors without requiring a fixed answer or solution path.

\paragraph{Process Supervision.}
Process supervision and PRMs provide denser feedback for math, code, and long-CoT reasoning, where intermediate correctness can often be assessed \cite{chen2024step,chen2025reasongrm,ye2025uncertainty,he2025good,zou2025reasonflux,fan2025posterior}. 
Such supervision is harder to transfer to open-ended alignment, which lacks unique solution paths and locally verifiable steps \cite{zhang2025openprm}. 
TCR addresses this setting by evaluating whole reasoning traces against preference-derived considerations rather than judging each step by local correctness. 
It also residualizes the checklist signal against outcome rewards, making the process signal complementary to final-answer supervision.

\paragraph{Rubric- and Checklist-based Evaluation.}
Rubric and checklist methods improve the granularity and interpretability of LLM evaluation and reward modeling \cite{lee2024checkeval,wang2025autorule,shen2026rethinking,liu2025openrubrics,zhou2026autochecklist,tian2026rit}.
These methods show that explicit natural-language criteria can capture preference-relevant dimensions beyond scalar scores, including for reasoning-oriented evaluation and supervision.
TCR is related to these studies in using explicit criteria, but differs by deriving sample-specific checklists from pairwise preferences and applying them to reasoning traces during online RL training. 
Moreover, TCR residualizes the checklist-based signal against the outcome reward to retain complementary process-level supervision beyond final-answer quality.

\section{Preliminary}
\label{sec:2}

\paragraph{Trajectory Generation}
Given a question $q$, a reasoning model $\mathcal{M}$ generates a response 
$o=(\mathcal{T},\mathcal{F})$, where $\mathcal{T}$ is the intermediate reasoning trace and $\mathcal{F}$ is the final answer.
The model first produces $\mathcal{T}$ and then generates $\mathcal{F}$ conditioned on both $q$ and $\mathcal{T}$, making the reasoning trajectory observable for supervision beyond the final answer.

\paragraph{Process Supervision}
Given $o=(\mathcal{T},\mathcal{F})$, outcome-level rewards assign a scalar reward $r^{out}$ mainly according to the quality of $\mathcal{F}$.
Process supervision instead provides an auxiliary signal for $\mathcal{T}$.
In open-ended alignment, step-level supervision is difficult because preferences often depend on user intent, constraints, trade-offs, factual relevance, and contextual appropriateness.
We therefore focus on trajectory-level process supervision, which evaluates the whole reasoning trace according to preference-implied considerations.

\begin{figure*}
    \centering
    \includegraphics[width=0.95\linewidth]{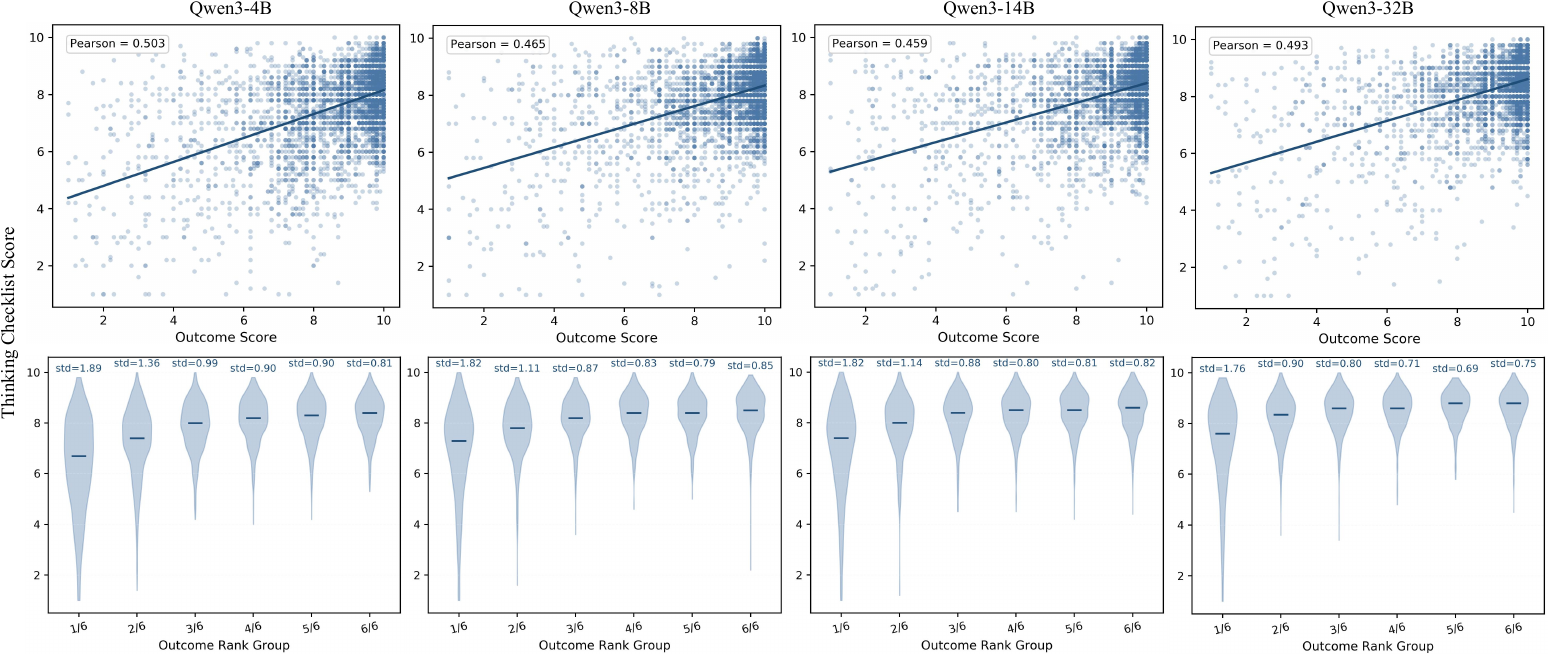}
\caption{Empirical relationship between outcome scores and thinking checklist scores. Top row: checklist scores are moderately positively correlated with outcome scores, showing that checklist-based evaluation is globally aligned with outcome-level quality. Bottom row: when samples are grouped by outcome rank, checklist scores still exhibit substantial within-group variation, indicating that they provide additional fine-grained resolution beyond outcome-level reward.}
    \label{fig:tcr}
\end{figure*}

\section{Thinking Checklist}
\label{sec:tc}
In this section, we define a \texttt{Thinking Checklist} as a sample-specific set of criteria for evaluating the reasoning trace $\mathcal{T}$. 
Each checklist is inferred from preference pairs and captures preference-implied considerations such as intent, constraints, trade-offs, and contextual appropriateness. 

\subsection{Empirical Analysis}
We next examine whether checklist scores provide a useful process-level signal. 
Such a signal should be broadly aligned with outcome quality while retaining local sensitivity beyond final-response scores. 
We use GPT-4o\footnote{\url{https://developers.openai.com/api/docs/models/gpt-4o}} to assign both outcome and checklist scores across Qwen3 models of different scales.

\paragraph{Coarse alignment with outcome quality.}
As shown in Figure~\ref{fig:tcr} (top row), checklist scores are moderately correlated with outcome scores across model scales, with Pearson correlations from 0.459 to 0.503. 
This moderate correlation is important: it indicates that checklist evaluation is not an arbitrary trajectory preference, but remains aligned with final-response quality at the global level. 
In other words, responses that receive higher outcome scores also tend to exhibit reasoning traces that better satisfy the corresponding preference-derived checklist.

\paragraph{Fine-grained sensitivity beyond outcome rewards.}
This correlation does not imply redundancy. 
Figure~\ref{fig:tcr} (bottom row) shows substantial checklist-score variation among responses with similar outcome ranks, especially in lower-outcome groups where trajectory quality is more diverse. 
Thus, outcome rewards can be too coarse for policy optimization: two responses may receive similar final-answer scores while differing in whether their reasoning considers the user's constraints, trade-offs, or contextual requirements. 
Checklist scores provide additional resolution over such trajectories. 
Together, the global correlation and within-rank variation motivate our residual design: instead of naively adding checklist rewards, we residualize them against outcome rewards and retain only the complementary thinking surplus for RL training.

\section{Method}
\label{sec:method}
\begin{figure*}
    \centering
    \includegraphics[width=0.95\linewidth]{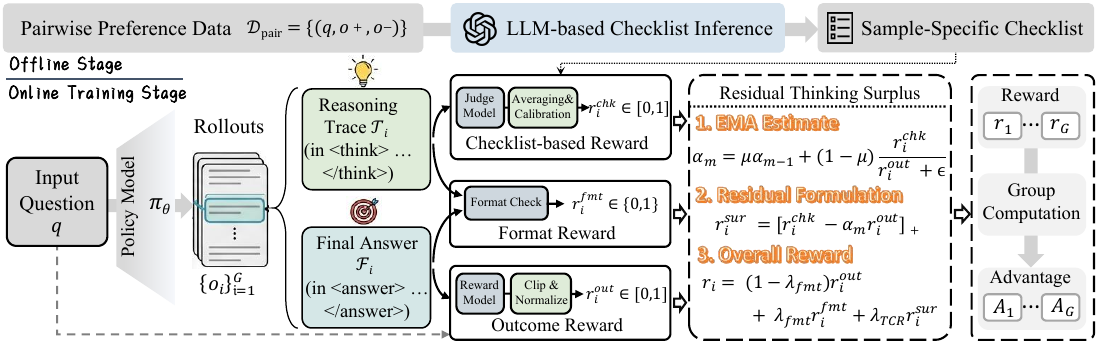}
\caption{Overview of the training pipeline with TCR. A sample-specific thinking checklist is first constructed from pairwise data, and the generated reasoning trace is evaluated against it to obtain a checklist-based reward. This reward is then residualized and combined with the outcome-level reward and format reward to produce the overall training reward for policy optimization.}
    \label{fig:method}
\end{figure*}
We present Thinking Checklist Reward (TCR), a process-oriented reward for open-ended alignment. As illustrated in Figure~\ref{fig:method}, TCR is constructed from sample-specific checklists inferred from pairwise preference data and used as an additional trajectory-level reward term within RL post-training. It is combined with an outcome reward and a format reward to form the scalar reward for policy optimization.

\subsection{Thinking Checklist Construction}

Let $\mathcal{D}_{\mathrm{pair}}=\{(q,o^{+},o^{-})\}$ denote pairwise preference data. For each pair, we use GPT-4o to infer a sample-specific checklist $\mathcal{C}(q)=\{c_k\}_{k=1}^{K_q}$, where each item represents a preference-implied consideration such as user intent, task constraints, trade-offs, or contextual appropriateness. This converts preference information into a trajectory-level supervision target without assuming locally verifiable reasoning steps.
The checklist is constructed offline before RL training and shared by all sampled responses for the same question. 

\subsection{Reward Formulation}

Given a sampled response $o_i=(\mathcal{T}_i,\mathcal{F}_i)$, we define three reward components: an outcome-level reward on the final answer, a format reward on the output structure, and a checklist-based reward on the reasoning trace.

\paragraph{(1) Outcome-level Reward $r_i^{\mathrm{out}}$.}
This reward evaluates the quality of the final answer $\mathcal{F}_i$ using a conventional reward model. Formally, we write $r_i^{\mathrm{out}}=\mathcal{R}_{\mathrm{out}}(q,\mathcal{F}_i)$. It serves as the main supervision signal during training and preserves the standard outcome-level alignment objective. In implementation, the raw score is clipped to a fixed interval and linearly normalized into $[0,1]$.

\paragraph{(2) Format Reward $r_i^{\mathrm{fmt}}$.}
This binary reward ensures structural compliance by verifying whether the model output follows the prescribed format, namely, placing the reasoning trace inside \texttt{<think>...</think>} and the final answer inside \texttt{<answer>...</answer>}. We assign $r_i^{\mathrm{fmt}}=1$ if the format is correct, and $r_i^{\mathrm{fmt}}=0$ otherwise. This term mainly supports reliable extraction of $\mathcal{T}_i$ and $\mathcal{F}_i$, a design that has also been adopted in prior work \cite{xie2025logic,guo2025deepseek}.

\paragraph{(3) Raw Checklist-based Reward $r_i^{\mathrm{chk}}$.}
To score the reasoning trace, we use a judge model, Qwen3-30B-A3B-Instruct-2507-FP8 \cite{qwen3technicalreport}, to evaluate how well $\mathcal{T}_i$ addresses the checklist items in $\mathcal{C}(q)$. Each checklist item is scored on a discrete five-level scale, with corresponding values in $\{2,4,6,8,10\}$. Let $s_k(\mathcal{T}_i,c_k)$ denote the score assigned to $\mathcal{T}_i$ on checklist item $c_k$. We first compute the overall checklist score by averaging the item-level scores:
\begin{equation}
\bar{s}_i
=
\frac{1}{K_q}\sum_{k=1}^{K_q}s_k(\mathcal{T}_i,c_k).
\end{equation}
We then transform it into a centered process reward,
\begin{equation}
r_i^{\mathrm{chk}}
=
\left[\frac{\bar{s}_i-6}{4}\right]_+,
\end{equation}
where $[\cdot]_+=\max(\cdot,0)$. This transformation serves two purposes. First, it centers the checklist score at the \emph{adequate} level, so that the reward reflects reasoning quality beyond a merely acceptable trace. Second, by retaining only the non-negative part, it lets the checklist-based term provide positive process supervision for sufficiently good reasoning without allowing low-confidence or noisy negative deviations to dominate the reward.

As motivated by the empirical analysis, directly adding $r_i^{\mathrm{chk}}$ would overlap with outcome-level supervision. We therefore transform it into TCR through a residual formulation.

\begin{table*}[t]
\centering
\footnotesize
\setlength{\tabcolsep}{2.5pt}
\renewcommand{\arraystretch}{0.88}
\begin{tabular}{lcc|ccc|ccc|ccc|r}
\toprule
\multirow{2}{*}{Base LLM} 
& \multicolumn{2}{c|}{Method}
& \multicolumn{3}{c|}{Vicuna Eval}
& \multicolumn{3}{c|}{Dolly Eval}
& \multicolumn{3}{c|}{BPO-test Eval}
& \multirow{2}{*}{$\Delta$WR} \\
\cmidrule(lr){2-3}\cmidrule(lr){4-6}\cmidrule(lr){7-9}\cmidrule(lr){10-12}
& A & B & A win & tie & B win & A win & tie & B win & A win & tie & B win & \\
\midrule
\multicolumn{13}{c}{\textit{\qwenicon\ Qwen Models}} \\
\midrule
\multirow{5}{*}{Qwen2.5-3B}
& DAPO & ori          & 50.00 & 2.50  & 47.50 & 47.50 & 15.00 & 37.50 & 46.00 & 16.50 & 37.50 & \textcolor{red}{+7.00} \\
& DPO          & ori          & 48.75 & 3.75  & 47.50 & 46.50 & 17.50 & 36.00 & 44.50 & 16.50 & 39.00 & \textcolor{red}{+5.75} \\
& DAPO + TCR    & DPO          & 58.75 & 2.50  & 38.75 & 44.00 & 17.50 & 38.50 & 44.00 & 18.50 & 37.50 & \textcolor{red}{+10.67} \\
& DAPO + TCR    & DAPO & 51.25 & 8.75  & 40.00 & 45.00 & 22.00 & 33.00 & 38.00 & 26.50 & 35.50 & \textcolor{red}{+8.58} \\
& DAPO + TCR    & w/o EMA      & 50.00 & 7.50  & 42.50 & 51.50 & 11.00 & 37.50 & 45.00 & 12.50 & 42.50 & \textcolor{red}{+8.00} \\
\midrule
\multirow{5}{*}{Qwen2.5-7B}
& DAPO & ori          & 50.00 & 8.75  & 41.25 & 51.00 & 18.50 & 30.50 & 43.50 & 21.00 & 35.50 & \textcolor{red}{+12.42} \\
& DPO          & ori          & 51.25 & 7.50  & 41.25 & 40.50 & 25.00 & 34.50 & 40.50 & 19.50 & 40.00 & \textcolor{red}{+5.50} \\
& DAPO + TCR    & DPO          & 46.25 & 8.75  & 45.00 & 58.00 & 13.50 & 28.50 & 50.00 & 16.50 & 33.50 & \textcolor{red}{+15.75} \\
& DAPO + TCR    & DAPO & 51.25 & 13.75 & 35.00 & 42.50 & 23.50 & 34.00 & 42.00 & 24.00 & 34.00 & \textcolor{red}{+10.92} \\
& DAPO + TCR    & w/o EMA      & 47.50 & 7.50  & 45.00 & 42.50 & 19.00 & 38.50 & 41.50 & 19.00 & 39.50 & \textcolor{red}{+2.83} \\
\midrule
\multicolumn{13}{c}{\textit{\llamaicon\ Llama Models}} \\
\midrule
\multirow{5}{*}{Llama3.2-3B}
& DAPO & ori          & 47.50 & 6.25  & 46.25 & 45.50 & 24.50 & 30.00 & 45.50 & 12.50 & 42.00 & \textcolor{red}{+6.75} \\
& DPO          & ori          & 53.75 & 3.75  & 42.50 & 45.00 & 23.50 & 31.50 & 46.50 & 15.50 & 38.00 & \textcolor{red}{+11.08} \\
& DAPO + TCR    & DPO          & 47.50 & 5.00  & 47.50 & 51.50 & 21.50 & 27.00 & 45.00 & 15.50 & 39.50 & \textcolor{red}{+10.00} \\
& DAPO + TCR    & DAPO & 51.25 & 8.75  & 40.00 & 42.50 & 23.50 & 34.00 & 47.50 & 13.00 & 39.50 & \textcolor{red}{+9.25} \\
& DAPO + TCR    & w/o EMA      & 53.75 & 5.00  & 41.25 & 39.50 & 28.00 & 32.50 & 48.00 & 11.00 & 41.00 & \textcolor{red}{+8.83} \\
\midrule
\multirow{5}{*}{Llama3.1-8B}
& DAPO & ori          & 55.00 & 5.00  & 40.00 & 45.00 & 22.50 & 32.50 & 46.50 & 15.00 & 38.50 & \textcolor{red}{+11.83} \\
& DPO          & ori          & 53.75 & 5.00  & 41.25 & 40.00 & 20.50 & 39.50 & 42.50 & 17.00 & 40.50 & \textcolor{red}{+5.00} \\
& DAPO + TCR    & DPO          & 52.50 & 7.50  & 40.00 & 49.50 & 17.50 & 33.00 & 55.50 & 9.00  & 35.50 & \textcolor{red}{+16.33} \\
& DAPO + TCR    & DAPO & 52.50 & 5.00  & 42.50 & 40.00 & 22.50 & 37.50 & 45.50 & 19.50 & 35.00 & \textcolor{red}{+7.67} \\
& DAPO + TCR    & w/o EMA      & 47.50 & 5.00  & 47.50 & 42.50 & 18.50 & 39.00 & 44.50 & 12.50 & 43.00 & \textcolor{red}{+1.67} \\
\midrule
\multicolumn{13}{c}{\textit{\deepseekicon\ DeepSeek Model}} \\
\midrule
\multirow{5}{*}{DeepSeek-llm-7B}
& DAPO & ori          & 48.75 & 11.25 & 40.00 & 41.00 & 23.50 & 35.50 & 42.50 & 18.00 & 39.50 & \textcolor{red}{+5.75} \\
& DPO          & ori          & 51.25 & 3.75  & 45.00 & 45.00 & 13.50 & 41.50 & 47.00 & 14.50 & 38.50 & \textcolor{red}{+6.08} \\
& DAPO + TCR    & DPO          & 56.25 & 5.00  & 38.75 & 44.00 & 13.50 & 42.50 & 49.50 & 9.50  & 41.00 & \textcolor{red}{+9.17} \\
& DAPO + TCR    & DAPO & 50.00 & 13.75 & 36.25 & 41.00 & 20.50 & 38.50 & 47.00 & 13.00 & 40.00 & \textcolor{red}{+7.75} \\
& DAPO + TCR    & w/o EMA      & 47.50 & 12.50 & 40.00 & 42.50 & 17.00 & 40.50 & 44.50 & 17.50 & 38.00 & \textcolor{red}{+5.33} \\
\bottomrule
\end{tabular}
\caption{Pairwise evaluation results on Vicuna Eval, Dolly Eval, and BPO-test Eval using GPT-4.1 as the judge model. $\Delta$WR is the average win-rate difference $(\text{A win}-\text{B win})$ across the three benchmarks. \textsc{DAPO+TCR} denotes DAPO training with the proposed TCR, and \textsc{w/o EMA} removes the EMA-based residualization.}
\label{tab:pairwise_main}
\end{table*}

\begin{table}[t]
\centering
\footnotesize
\setlength{\tabcolsep}{3.6pt}
\renewcommand{\arraystretch}{1.08}
\resizebox{\columnwidth}{!}{
\begin{tabular}{ll|ccccc}
\toprule
Base LLM & Metric & ori & DAPO & DPO & DAPO + TCR & w/o EMA \\
\midrule
\multicolumn{7}{c}{\textit{\qwenicon\ Qwen Models}} \\
\midrule
\multirow{3}{*}{Qwen2.5-3B}
& WR   & 12.99 & 14.48 & 12.39 & \textbf{14.80} & 12.68 \\
& DWR  & 12.92 & 14.41 & 12.30 & \textbf{14.97} & 12.61 \\
& LCWR & 11.75 & 12.74 & 11.92 & \textbf{13.44} & 11.47 \\
\midrule
\multirow{3}{*}{Qwen2.5-7B}
& WR   & 23.88 & 24.62 & 22.63 & \textbf{27.14} & 26.43 \\
& DWR  & 24.65 & 24.09 & 23.47 & \textbf{27.19} & 26.46 \\
& LCWR & 23.83 & 24.05 & 24.64 & \textbf{25.79} & 24.49 \\
\midrule
\multicolumn{7}{c}{\textit{\llamaicon\ Llama Models}} \\
\midrule
\multirow{3}{*}{Llama3.2-3B}
& WR   & 11.31 & 11.92 & 13.06 & \textbf{13.84} & 13.59 \\
& DWR  & 11.12 & 11.86 & 13.04 & \textbf{13.66} & 13.54 \\
& LCWR & 10.09 & 11.13 & 11.62 & \textbf{13.10} & 12.67 \\
\midrule
\multirow{3}{*}{Llama3.1-8B}
& WR   & 17.28 & 19.83 & 18.08 & \textbf{20.03} & 19.34 \\
& DWR  & 17.20 & 19.94 & 18.39 & \textbf{20.06} & 19.01 \\
& LCWR & 15.48 & 16.85 & 16.82 & \textbf{16.87} & 16.20 \\
\midrule
\multicolumn{7}{c}{\textit{\deepseekicon\ DeepSeek Model}} \\
\midrule
\multirow{3}{*}{DeepSeek-llm-7B}
& WR   & 3.44 & 3.06 & 3.40 & \textbf{4.25} & 3.36 \\
& DWR  & 3.42 & 2.98 & 3.35 & \textbf{4.16} & 3.23 \\
& LCWR & 4.74 & 4.28 & 5.52 & \textbf{5.96} & 4.85 \\
\bottomrule
\end{tabular}
}
\caption{
Results on AlpacaEval~2.0. 
WR, DWR, and LCWR denote \texttt{win\_rate}, \texttt{discrete\_win\_rate}, and \texttt{length\_controlled\_winrate}, respectively.
}
\label{tab:alpacaeval_compact}
\end{table}
\subsection{Residual Thinking Surplus}

We maintain a global exponential moving average that is updated sequentially over sampled responses during reward computation.
Let $m$ denote the index of the response-level reward computation. 
For the sampled response $o_i$, we update
\begin{equation}
\alpha_m
=
\mu \alpha_{m-1}
+
(1-\mu)
\frac{r_i^{\mathrm{chk}}}{r_i^{\mathrm{out}}+\epsilon},
\end{equation}
where $\mu\in[0,1)$ is the decay factor and $\epsilon>0$ is a small constant for numerical stability. Intuitively, the ratio $r_i^{\mathrm{chk}}/(r_i^{\mathrm{out}}+\epsilon)$ characterizes the typical scale relationship between the raw checklist-based reward and the outcome-level reward, while $\alpha_m$ provides a stable running estimate of this relationship throughout training.
Based on this estimate, we define the residual thinking surplus as
\begin{equation}
r_i^{\mathrm{sur}}
=
\left[
r_i^{\mathrm{chk}}-\alpha_m r_i^{\mathrm{out}}
\right]_+,
\end{equation}
where $[\cdot]_+=\max(\cdot,0)$. Here, the subtraction term $\alpha_m r_i^{\mathrm{out}}$ represents the coarse component of checklist quality that is already predictable from final-response quality. Accordingly, $r_i^{\mathrm{sur}}$ retains only the additional part of reasoning quality that is not already captured by the final answer. The positive-part operator further ensures that this term acts as a complementary bonus, rather than introducing unnecessary penalties when the checklist-based reward is already well explained by the outcome-level reward.
We then define the TCR as
\begin{equation}
r_i^{\mathrm{TCR}}
=
\lambda_{\mathrm{TCR}}\,r_i^{\mathrm{sur}},
\end{equation}
where $\lambda_{\mathrm{TCR}}$ controls the strength of process-level supervision. In this way, TCR is not the overall training reward itself, but the residualized checklist-based reward term introduced to complement conventional outcome-level supervision.

\subsection{Overall Training Reward}

We now combine the above components into the final scalar reward used for RL training:
\begin{equation}
r_i
=
(1-\lambda_{\mathrm{fmt}})\,r_i^{\mathrm{out}}
+
\lambda_{\mathrm{fmt}}\,r_i^{\mathrm{fmt}}
+
r_i^{\mathrm{TCR}},
\end{equation}
where $\lambda_{\mathrm{fmt}}$ controls the contribution of the format reward. The first term anchors optimization to the conventional outcome-level objective, the second term encourages structured outputs, and the third term injects additional supervision on the reasoning trace through the residualized checklist-based reward. In this way, the final training reward preserves the original alignment direction while incorporating complementary process-level information.

\subsection{Policy Optimization}

We optimize the policy $\pi_{\theta}$ using the DAPO framework \cite{yu2025dapo}, where $\theta$ denotes the policy parameters and $r_i$ is the overall reward defined above. For each input question $q$, the behavior policy $\pi_{\theta_{\mathrm{old}}}$ samples a group of responses
\[
\{o_i\}_{i=1}^{G}\sim\pi_{\theta_{\mathrm{old}}}(\cdot\mid q).
\]
Their rewards are then normalized within the group to obtain relative advantages:
\begin{equation}
\hat{A}_i
=
\frac{
r_i-\mathrm{mean}(\{r_j\}_{j=1}^{G})
}{
\mathrm{std}(\{r_j\}_{j=1}^{G})+\delta
},
\end{equation}
where $\delta$ is a small constant for numerical stability. The resulting advantage is shared across all tokens in the same response.

For the $t$-th token of response $o_i$, the importance ratio is defined as
\begin{equation}
\rho_{i,t}(\theta)
=
\frac{
\pi_{\theta}(o_{i,t}\mid q,o_{i,<t})
}{
\pi_{\theta_{\mathrm{old}}}(o_{i,t}\mid q,o_{i,<t})
},
\end{equation}
and its decoupled clipped counterpart is given by
\begin{equation}
\bar{\rho}_{i,t}(\theta)
=
\mathrm{clip}\!\left(
\rho_{i,t}(\theta),
1-\varepsilon_{\mathrm{low}},
1+\varepsilon_{\mathrm{high}}
\right),
\end{equation}
where $\varepsilon_{\mathrm{low}}$ and $\varepsilon_{\mathrm{high}}$ are the lower and upper clipping coefficients, respectively. Based on these two quantities, the token-level clipped surrogate is then
\begin{equation}
\ell_{i,t}(\theta)
=
\min\!\Big(
\rho_{i,t}(\theta)\hat{A}_i,\,
\bar{\rho}_{i,t}(\theta)\hat{A}_i
\Big).
\end{equation}

The optimization objective is given by
\begin{equation}
\mathcal{J}(\theta)
=
\mathbb{E}\!\left[
\frac{1}{\sum_{i=1}^{G}|o_i|}
\sum_{i=1}^{G}
\sum_{t=1}^{|o_i|}
\ell_{i,t}(\theta)
\right],
\end{equation}
where the expectation is taken over $q$ and $\{o_i\}_{i=1}^{G}\sim\pi_{\theta_{\mathrm{old}}}(\cdot\mid q)$.

Overall, policy optimization in our method follows the DAPO framework with GRPO-style group-normalized advantages and decoupled clipping, while omitting the KL term and disabling dynamic sampling in our experiments.

\section{Experiments}
\subsection{Experiment Setups}

\paragraph{Evaluation.}
We evaluate our method on four open-ended benchmarks: Dolly Eval \cite{conover2023free}, Vicuna Eval \cite{chiang2023vicuna}, BPO-test Eval \cite{cheng2024black}, and AlpacaEval~2.0 \cite{li2023alpacaeval}. 
For all benchmarks, we adopt a pairwise evaluation protocol using GPT-4.1\footnote{\url{https://openai.com/index/gpt-4-1}} as the judge model. 
The evaluation prompt is based on the AlpacaEval scoring template. 
To mitigate position bias, we randomly shuffle the order of model responses in each comparison. 
Additional dataset and prompt details are provided in the supplementary material.

\paragraph{Training.}
We conduct experiments on five LLMs from three model families: Qwen2.5-3B, Qwen2.5-7B \cite{qwen2.5}, Llama3.2-3B, Llama3.1-8B \cite{grattafiori2024llama}, and DeepSeek-LLM-7B-chat \cite{bi2024deepseek}. 
All models are trained on the BPO dataset \cite{cheng2024black} using the \texttt{verl} framework \cite{sheng2025hybridflow}. 
Our method follows the DAPO optimization and incorporates TCR as an additional residualized checklist-based reward term. 
The thinking checklists are constructed offline from pairwise preference data and used only during reward computation. 
Additional hyperparameters are provided in the supplementary material.

\paragraph{Baselines.}
For each model, we compare against three baselines: the original base LLM, DPO, and DAPO. 
The base LLM measures the effect of post-training, while DPO serves as an offline preference optimization baseline using the same type of pairwise supervision. 
DAPO uses the same RL optimization framework as our method but excludes TCR, allowing us to directly evaluate the contribution of the proposed TCR.

 \subsection{Main Results}

Tables~\ref{tab:pairwise_main} and~\ref{tab:alpacaeval_compact} present the main results on open-ended alignment benchmarks. 
Overall, incorporating TCR into DAPO consistently improves alignment performance across model families and evaluation settings.
As shown in Table~\ref{tab:pairwise_main}, DAPO+TCR achieves positive average win-rate differences over DAPO on all five backbone models across Vicuna Eval, Dolly Eval, and BPO-test Eval, with gains of +8.58, +10.92, +9.25, +7.67, and +7.75 $\Delta$WR, respectively. 
Since DAPO+TCR and DAPO use the same RL optimization framework, these improvements indicate that the gains come from the additional trajectory-level reward introduced by TCR. 
DAPO+TCR also generally outperforms DPO, suggesting that converting pairwise preference information into checklist-based reasoning supervision provides benefits beyond standard offline preference optimization.
Table~\ref{tab:alpacaeval_compact} further shows that DAPO+TCR achieves the best results on AlpacaEval~2.0 across all five backbones and all three metrics, including WR, DWR, and LCWR. 
The consistent improvement in LCWR suggests that the gains are not merely due to longer responses, but reflect better length-controlled alignment quality. 
Together, these results show that TCR is an effective reward component for improving DAPO-based open-ended RL alignment.
Bootstrap confidence intervals over pooled instances, reported in the supplementary material, further support the observed improvements.

\begin{figure}[t]
    \centering
    \includegraphics[width=0.9\columnwidth]{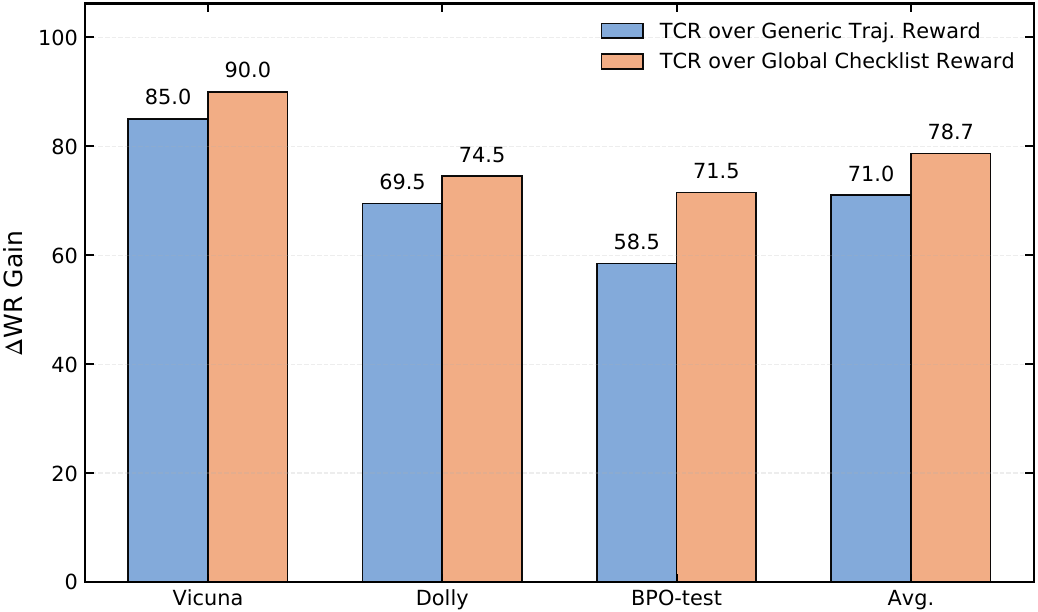}
    \caption{
Ablation of reward design on Qwen2.5-7B.
We report the gain of TCR over Generic Traj. Reward and Global Checklist Reward.
}
    \label{fig:reward_design_gain}
\end{figure}

\subsection{Ablation Study}

We conduct ablations on the key design choices of TCR.

\paragraph{Effect of EMA-based residual formulation.}
To examine the role of residualization, we compare TCR with a w/o EMA variant that directly uses the raw checklist-based reward.
As shown in Table~\ref{tab:pairwise_main}, DAPO+TCR consistently outperforms this variant across all five backbones, with average gains of +8.00, +2.83, +8.83, +1.67, and +5.33 $\Delta$WR, respectively.
The same trend is observed on AlpacaEval~2.0 in Table~\ref{tab:alpacaeval_compact}.
These results show that the EMA-based residual formulation is more effective than directly adding raw checklist rewards.
This supports our motivation that checklist rewards should be dynamically calibrated against outcome rewards, rather than treated as an independent additive signal.

\paragraph{Reward design comparison.}
We further compare TCR with two controlled variants on Qwen2.5-7B.
Generic Traj. Reward directly scores the whole reasoning trajectory, while Global Checklist Reward replaces the sample-specific checklist with a unified checklist shared across samples; prompt details are provided in the supplementary material.
As shown in Figure~\ref{fig:reward_design_gain}, TCR achieves clear $\Delta$WR gains over both variants across Vicuna, Dolly, and BPO-test.
This suggests that TCR's improvement is not merely from adding a generic trajectory-level reward; preference-derived sample-specific checklist criteria provide more targeted process supervision.

\begin{figure}[t]
    \centering
    \includegraphics[width=\columnwidth]{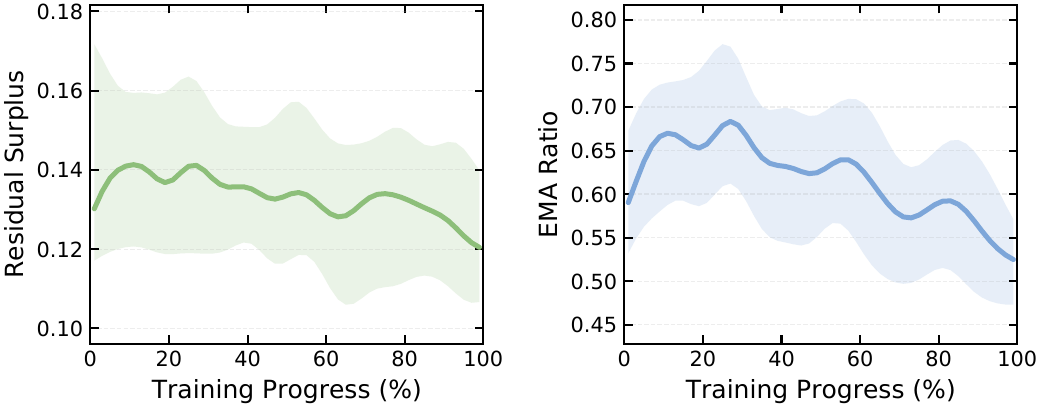}
    \caption{
    Reward component analysis on Qwen2.5-7B.
    Left: residual surplus $r_{\mathrm{sur}}$.
    Right: EMA ratio.
    }
    \label{fig:reward_component}
\end{figure}
\subsection{Additional Analysis}

\paragraph{Reward components.}
To better understand how residualization behaves during training, we track the residual surplus and EMA ratio on Qwen2.5-7B.
As shown in Figure~\ref{fig:reward_component} (left), the residual surplus remains consistently above zero, indicating that the checklist signal is not fully absorbed by the outcome reward and continues to provide auxiliary process-level supervision throughout training.
Figure~\ref{fig:reward_component} (right) shows that the EMA ratio gradually decreases during training.
This is expected because the overall reward is dominated by the outcome reward, so policy optimization primarily improves final-answer quality, while checklist satisfaction need not increase at the same rate.
The changing ratio also suggests that the coupling between checklist and outcome rewards is not fixed, supporting the use of EMA-based adaptive calibration.


\begin{figure}[t]
    \centering
    \includegraphics[width=\columnwidth]{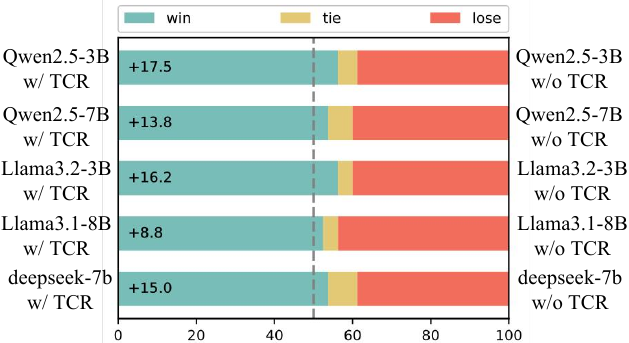}
    \caption{Vicuna Eval's evaluation using Gemini 2.5 Pro.}
    \label{fig:judge_robustness}
\end{figure}


\paragraph{Evaluator robustness.}
Since our main evaluation relies on GPT-4.1, we further test whether the observed gains are robust to the choice of judge model.
Specifically, we evaluate DAPO+TCR against DAPO on Vicuna Eval using Gemini-2.5-Pro\footnote{\url{https://ai.google.dev/gemini-api/docs/models/gemini-2.5-pro}} as an additional judge.
As shown in Figure~\ref{fig:judge_robustness}, Gemini also prefers DAPO+TCR over DAPO across all five backbones.
This suggests that the improvement is not specific to GPT-4.1 and is less likely to be an artifact of a single evaluator.

\begin{figure}[t]
    \centering
    \includegraphics[width=0.9\columnwidth]{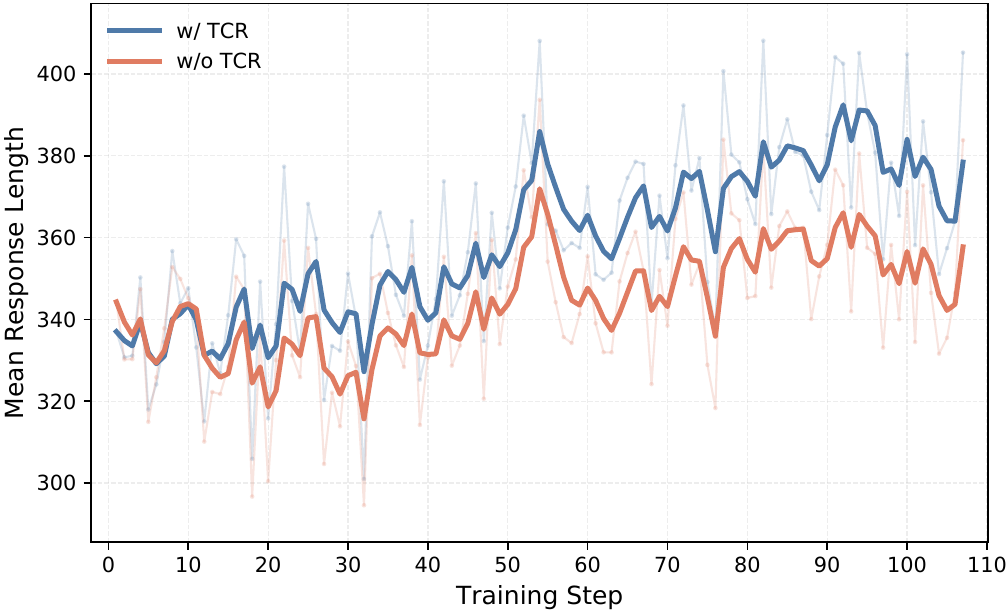}
    \caption{Mean response length during Qwen2.5-7B training.}
    \label{fig:response_length}
\end{figure}

\begin{figure}[t]
    \centering
    \includegraphics[width=0.9\columnwidth]{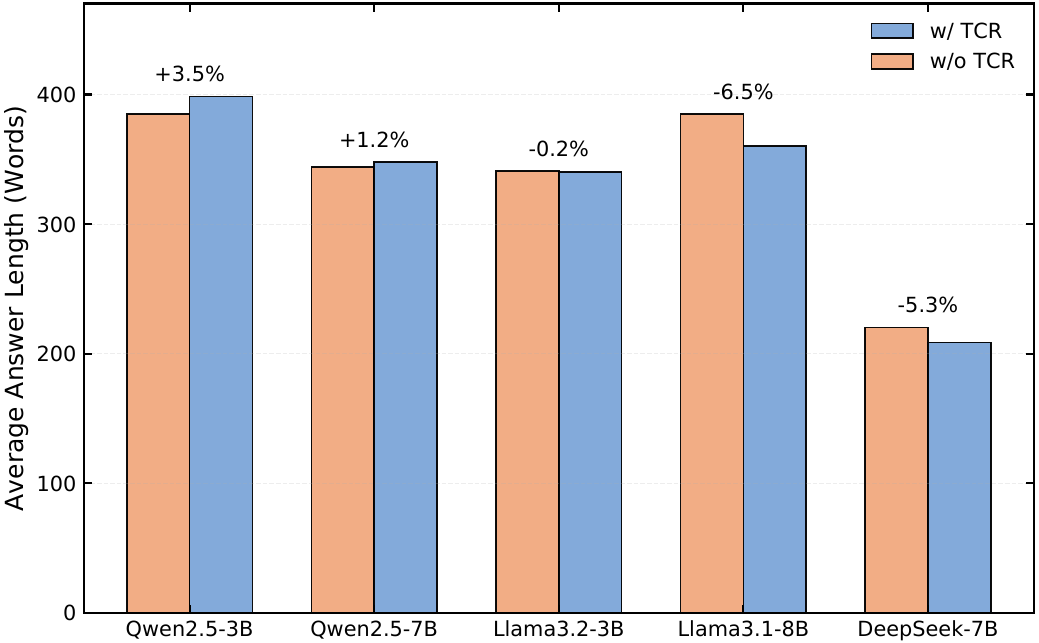}
    \caption{Average answer length on AlpacaEval~2.0.}
    \label{fig:answer_length}
\end{figure}
\paragraph{Response length.}
We examine whether TCR's gains mainly come from longer generations.
Figure~\ref{fig:response_length} shows that DAPO+TCR produces longer overall responses during Qwen2.5-7B training, indicating a change in generation behavior.
Yet Figure~\ref{fig:answer_length} shows that final answer length remains largely unchanged on AlpacaEval~2.0, suggesting that this change mainly occurs in the reasoning rather than final answers.
Together with the LCWR improvements in Table~\ref{tab:alpacaeval_compact}, this suggests that TCR improves alignment quality by shaping the thinking process, without relying on answer-length inflation.

\section{Conclusion}

We propose TCR, a process-oriented reward for open-ended LLM preference alignment. 
TCR converts preference information into sample-specific checklists to evaluate reasoning traces, providing process supervision beyond outcome-level rewards. 
To reduce reward overlap, TCR introduces an EMA-based residual formulation that retains complementary thinking surplus beyond final-response reward. 
Experiments on five LLMs from three families show that adding TCR to DAPO consistently improves alignment performance over DAPO and DPO baselines. 
Ablation, length analysis, and evaluator-robustness results further support the effectiveness of the proposed reward design, with additional checklist analyses and case studies provided in the supplementary material.

\bibliography{aaai2027}

@article{yang2025qwen3,
  title={Qwen3 technical report},
  author={Yang, An and Li, Anfeng and Yang, Baosong and Zhang, Beichen and Hui, Binyuan and Zheng, Bo and Yu, Bowen and Gao, Chang and Huang, Chengen and Lv, Chenxu and others},
  journal={arXiv preprint arXiv:2505.09388},
  year={2025}
}

@article{guo2025deepseek,
  title={Deepseek-r1: Incentivizing reasoning capability in llms via reinforcement learning},
  author={Guo, Daya and Yang, Dejian and Zhang, Haowei and Song, Junxiao and Wang, Peiyi and Zhu, Qihao and Xu, Runxin and Zhang, Ruoyu and Ma, Shirong and Bi, Xiao and others},
  journal={arXiv preprint arXiv:2501.12948},
  year={2025}
}

@article{shao2024deepseekmath,
  title={Deepseekmath: Pushing the limits of mathematical reasoning in open language models},
  author={Shao, Zhihong and Wang, Peiyi and Zhu, Qihao and Xu, Runxin and Song, Junxiao and Bi, Xiao and Zhang, Haowei and Zhang, Mingchuan and Li, YK and Wu, Yang and others},
  journal={arXiv preprint arXiv:2402.03300},
  year={2024}
}

@article{zhang2025auditable,
  title={Auditable-choice reframing unlocks RL-based verification for open-ended tasks},
  author={Zhang, Mengyu and Liu, Xubo and Ding, Siyu and Yin, Weichong and Sun, Yu and Wu, Hua and Guo, Wenya and Zhang, Ying},
  journal={arXiv e-prints},
  pages={arXiv--2511},
  year={2025}
}

@article{simonds2025rlsr,
  title={RLSR: Reinforcement Learning from Self Reward},
  author={Simonds, Toby and Lopez, Kevin and Yoshiyama, Akira and Garmier, Dominique},
  journal={arXiv preprint arXiv:2505.08827},
  year={2025}
}

@article{shen2026rethinking,
  title={Rethinking Rubric Generation for Improving LLM Judge and Reward Modeling for Open-ended Tasks},
  author={Shen, William F and Qiu, Xinchi and Whitehouse, Chenxi and Alazraki, Lisa and Goel, Shashwat and Barbieri, Francesco and Willi, Timon and Mathur, Akhil and Leontiadis, Ilias},
  journal={arXiv preprint arXiv:2602.05125},
  year={2026}
}

@article{su2025crossing,
  title={Crossing the reward bridge: Expanding rl with verifiable rewards across diverse domains},
  author={Su, Yi and Yu, Dian and Song, Linfeng and Li, Juntao and Mi, Haitao and Tu, Zhaopeng and Zhang, Min and Yu, Dong},
  journal={arXiv preprint arXiv:2503.23829},
  year={2025}
}

@article{xu2025direct,
  title={Direct reasoning optimization: Llms can reward and refine their own reasoning for open-ended tasks},
  author={Xu, Yifei and Chakraborty, Tusher and Sharma, Srinagesh and Nunes, Leonardo and K{\i}c{\i}man, Emre and Lu, Songwu and Chandra, Ranveer},
  journal={arXiv e-prints},
  pages={arXiv--2506},
  year={2025}
}

@article{liu2024skywork,
  title={Skywork-Reward: Bag of Tricks for Reward Modeling in LLMs},
  author={Liu, Chris Yuhao and Zeng, Liang and Liu, Jiacai and Yan, Rui and He, Jujie and Wang, Chaojie and Yan, Shuicheng and Liu, Yang and Zhou, Yahui},
  journal={arXiv preprint arXiv:2410.18451},
  year={2024}
}

@article{chen2025reasongrm,
  title={Reasongrm: Enhancing generative reward models through large reasoning models},
  author={Chen, Bin and Gao, Xinzge and Hu, Chuanrui and Yu, Penghang and Zhang, Hua and Bao, Bing-Kun},
  journal={arXiv preprint arXiv:2506.16712},
  year={2025}
}

@inproceedings{chen2024step,
  title={Step-level value preference optimization for mathematical reasoning},
  author={Chen, Guoxin and Liao, Minpeng and Li, Chengxi and Fan, Kai},
  booktitle={Findings of the Association for Computational Linguistics: EMNLP 2024},
  pages={7889--7903},
  year={2024}
}

@article{zou2025reasonflux,
  title={Reasonflux-prm: Trajectory-aware prms for long chain-of-thought reasoning in llms},
  author={Zou, Jiaru and Yang, Ling and Gu, Jingwen and Qiu, Jiahao and Shen, Ke and He, Jingrui and Wang, Mengdi},
  journal={arXiv preprint arXiv:2506.18896},
  year={2025}
}

@article{ye2025uncertainty,
  title={Uncertainty-aware step-wise verification with generative reward models},
  author={Ye, Zihuiwen and Melo, Luckeciano Carvalho and Kaddar, Younesse and Blunsom, Phil and Staton, Sam and Gal, Yarin},
  journal={arXiv preprint arXiv:2502.11250},
  year={2025}
}

@article{he2025good,
  title={Good learners think their thinking: Generative PRM makes large reasoning model more efficient math learner},
  author={He, Tao and Mu, Rongchuan and Liao, Lizi and Cao, Yixin and Liu, Ming and Qin, Bing},
  journal={arXiv preprint arXiv:2507.23317},
  year={2025}
}

@article{fan2025posterior,
  title={Posterior-grpo: Rewarding reasoning processes in code generation},
  author={Fan, Lishui and Zhang, Yu and Chen, Mouxiang and Liu, Zhongxin},
  journal={arXiv preprint arXiv:2508.05170},
  year={2025}
}

@article{wang2025autorule,
  title={Autorule: Reasoning chain-of-thought extracted rule-based rewards improve preference learning},
  author={Wang, Tevin and Xiong, Chenyan},
  journal={arXiv preprint arXiv:2506.15651},
  year={2025}
}

@article{zheng2023judging,
  title={Judging llm-as-a-judge with mt-bench and chatbot arena},
  author={Zheng, Lianmin and Chiang, Wei-Lin and Sheng, Ying and Zhuang, Siyuan and Wu, Zhanghao and Zhuang, Yonghao and Lin, Zi and Li, Zhuohan and Li, Dacheng and Xing, Eric and others},
  journal={Advances in neural information processing systems},
  volume={36},
  pages={46595--46623},
  year={2023}
}

@inproceedings{cheng2024black,
  title={Black-box prompt optimization: Aligning large language models without model training},
  author={Cheng, Jiale and Liu, Xiao and Zheng, Kehan and Ke, Pei and Wang, Hongning and Dong, Yuxiao and Tang, Jie and Huang, Minlie},
  booktitle={Proceedings of the 62nd Annual Meeting of the Association for Computational Linguistics (Volume 1: Long Papers)},
  pages={3201--3219},
  year={2024}
}

@article{xie2025logic,
  title={Logic-rl: Unleashing llm reasoning with rule-based reinforcement learning},
  author={Xie, Tian and Gao, Zitian and Ren, Qingnan and Luo, Haoming and Hong, Yuqian and Dai, Bryan and Zhou, Joey and Qiu, Kai and Wu, Zhirong and Luo, Chong},
  journal={arXiv preprint arXiv:2502.14768},
  year={2025}
}

@misc{qwen3technicalreport,
      title={Qwen3 Technical Report}, 
      author={Qwen Team},
      year={2025},
      eprint={2505.09388},
      archivePrefix={arXiv},
      primaryClass={cs.CL},
      url={https://arxiv.org/abs/2505.09388}, 
}

@article{yu2025dapo,
  title={Dapo: An open-source llm reinforcement learning system at scale},
  author={Yu, Qiying and Zhang, Zheng and Zhu, Ruofei and Yuan, Yufeng and Zuo, Xiaochen and Yue, Yu and Dai, Weinan and Fan, Tiantian and Liu, Gaohong and Liu, Lingjun and others},
  journal={arXiv preprint arXiv:2503.14476},
  year={2025}
}

@article{conover2023free,
  title={Free dolly: Introducing the world’s first truly open instructiontuned llm},
  author={Conover, Mike and Hayes, Matt and Mathur, Ankit and Xie, Jianwei and Wan, Jun and Shah, Sam and Ghodsi, Ali and Wendell, Patrick and Zaharia, Matei and Xin, Reynold},
  year={2023}
}

@article{chiang2023vicuna,
  title={Vicuna: An open-source chatbot impressing gpt-4 with 90\%* chatgpt quality},
  author={Chiang, Wei-Lin and Li, Zhuohan and Lin, Ziqing and Sheng, Ying and Wu, Zhanghao and Zhang, Hao and Zheng, Lianmin and Zhuang, Siyuan and Zhuang, Yonghao and Gonzalez, Joseph E and others},
  journal={See https://vicuna. lmsys. org (accessed 14 April 2023)},
  volume={2},
  number={3},
  pages={6},
  year={2023}
}

@misc{li2023alpacaeval,
  title={Alpacaeval: An automatic evaluator of instruction-following models},
  author={Li, Xuechen and Zhang, Tianyi and Dubois, Yann and Taori, Rohan and Gulrajani, Ishaan and Guestrin, Carlos and Liang, Percy and Hashimoto, Tatsunori B},
  year={2023}
}

@inproceedings{sheng2025hybridflow,
  title={Hybridflow: A flexible and efficient rlhf framework},
  author={Sheng, Guangming and Zhang, Chi and Ye, Zilingfeng and Wu, Xibin and Zhang, Wang and Zhang, Ru and Peng, Yanghua and Lin, Haibin and Wu, Chuan},
  booktitle={Proceedings of the Twentieth European Conference on Computer Systems},
  pages={1279--1297},
  year={2025}
}

@misc{qwen2.5,
    title = {Qwen2.5: A Party of Foundation Models},
    url = {https://qwenlm.github.io/blog/qwen2.5/},
    author = {Qwen Team},
    month = {September},
    year = {2024}
}

@article{grattafiori2024llama,
  title={The llama 3 herd of models},
  author={Grattafiori, Aaron and Dubey, Abhimanyu and Jauhri, Abhinav and Pandey, Abhinav and Kadian, Abhishek and Al-Dahle, Ahmad and Letman, Aiesha and Mathur, Akhil and Schelten, Alan and Vaughan, Alex and others},
  journal={arXiv preprint arXiv:2407.21783},
  year={2024}
}

@article{bi2024deepseek,
  title={Deepseek llm: Scaling open-source language models with longtermism},
  author={Bi, Xiao and Chen, Deli and Chen, Guanting and Chen, Shanhuang and Dai, Damai and Deng, Chengqi and Ding, Honghui and Dong, Kai and Du, Qiushi and Fu, Zhe and others},
  journal={arXiv preprint arXiv:2401.02954},
  year={2024}
}

@article{kopf2023openassistant,
  title={Openassistant conversations-democratizing large language model alignment},
  author={K{\"o}pf, Andreas and Kilcher, Yannic and Von R{\"u}tte, Dimitri and Anagnostidis, Sotiris and Tam, Zhi Rui and Stevens, Keith and Barhoum, Abdullah and Nguyen, Duc and Stanley, Oliver and Nagyfi, Rich{\'a}rd and others},
  journal={Advances in neural information processing systems},
  volume={36},
  pages={47669--47681},
  year={2023}
}

@article{bai2022training,
  title={Training a helpful and harmless assistant with reinforcement learning from human feedback},
  author={Bai, Yuntao and Jones, Andy and Ndousse, Kamal and Askell, Amanda and Chen, Anna and DasSarma, Nova and Drain, Dawn and Fort, Stanislav and Ganguli, Deep and Henighan, Tom and others},
  journal={arXiv preprint arXiv:2204.05862},
  year={2022}
}

@article{peng2023instruction,
  title={Instruction tuning with gpt-4},
  author={Peng, Baolin and Li, Chunyuan and He, Pengcheng and Galley, Michel and Gao, Jianfeng},
  journal={arXiv preprint arXiv:2304.03277},
  year={2023}
}

@article{achiam2023gpt,
  title={Gpt-4 technical report},
  author={Achiam, Josh and Adler, Steven and Agarwal, Sandhini and Ahmad, Lama and Akkaya, Ilge and Aleman, Florencia Leoni and Almeida, Diogo and Altenschmidt, Janko and Altman, Sam and Anadkat, Shyamal and others},
  journal={arXiv preprint arXiv:2303.08774},
  year={2023}
}

@article{zhong2026p2s,
  title={P2S: Probabilistic Process Supervision for General-Domain Reasoning Question Answering},
  author={Zhong, Wenlin and Liu, Chengyuan and Wu, Yiquan and Tan, Bovin and Sun, Changlong and Wang, Yi and Liu, Xiaozhong and Kuang, Kun},
  journal={arXiv preprint arXiv:2601.20649},
  year={2026}
}

@inproceedings{zhang2025openprm,
  title={Openprm: Building open-domain process-based reward models with preference trees},
  author={Zhang, Kaiyan and Zhang, Jiayuan and Li, Haoxin and Zhu, Xuekai and Hua, Ermo and Lv, Xingtai and Ding, Ning and Qi, Biqing and Zhou, Bowen},
  booktitle={The thirteenth international conference on learning representations},
  year={2025}
}

@article{liu2025openrubrics,
  title={Openrubrics: Towards scalable synthetic rubric generation for reward modeling and llm alignment},
  author={Liu, Tianci and Xu, Ran and Yu, Tony and Hong, Ilgee and Yang, Carl and Zhao, Tuo and Wang, Haoyu},
  journal={arXiv preprint arXiv:2510.07743},
  year={2025}
}

@article{zhou2026autochecklist,
  title={AutoChecklist: Composable Pipelines for Checklist Generation and Scoring with LLM-as-a-Judge},
  author={Zhou, Karen and Tan, Chenhao},
  journal={arXiv preprint arXiv:2603.07019},
  year={2026}
}

@article{lee2024checkeval,
  title={Checkeval: Robust evaluation framework using large language model via checklist},
  author={Lee, Yukyung and Kim, Joonghoon and Kim, Jaehee and Cho, Hyowon and Kang, Pilsung},
  journal={arXiv preprint arXiv:2403.18771},
  volume={3},
  number={5},
  year={2024}
}

@inproceedings{ou2026serl,
  title={Serl: Self-examining reinforcement learning on open-domain},
  author={Ou, Weixuan and Zheng, Yanzhao and Sun, Shuoshuo and Zhang, Wei and Dong, Baohua and Zhu, Hangcheng and Huang, Ruohui and Yu, Gang and Yan, Pengwei and Qiao, Yifan},
  booktitle={Proceedings of the AAAI Conference on Artificial Intelligence},
  volume={40},
  number={38},
  pages={32619--32627},
  year={2026}
}

@inproceedings{tian2026rit,
  title={RiT: Rubrics-in-Thinking Reinforcement Learning for Improved Reasoning in Large Language Models},
  author={Tian, Xiaobin and Yuan, Shuai and Ding, Muyun and Chen, Haonan and Jiang, Xiaoxi},
  booktitle={Findings of the Association for Computational Linguistics: ACL 2026},
  pages={3944--3957},
  year={2026}
}

\appendix

\section{Reward Computation Algorithm}
\label{app:reward_algorithm}

Algorithm~\ref{alg:reward_computation} summarizes the overall reward computation with TCR during RL training.

\begin{algorithm}[!t]
\small
\caption{Reward Computation with TCR}
\label{alg:reward_computation}
\begin{algorithmic}[1]
\Require Output $o_i$, question $q$, checklist $\mathcal{C}(q)$
\Require Outcome RM $\mathcal{R}_{\mathrm{out}}$, checklist judge $\mathcal{J}_{\mathrm{chk}}$
\Require EMA $\alpha$, decay $\mu$, ratio bound $M$
\Require Weights $\lambda_{\mathrm{fmt}}, \lambda_{\mathrm{TCR}}$
\Ensure Final reward $r_i$

\State Extract $\mathcal{F}_i$ from \texttt{<answer>...</answer>}
\If{$\mathcal{F}_i$ is missing}
    \State \Return invalid reward
\EndIf
\State Extract $\mathcal{T}_i$ from \texttt{<think>...</think>}
\State Compute format reward:
\[
r_i^{\mathrm{fmt}}
=
\mathbb{I}\{o_i \text{ has valid tags}\}.
\]
\State Compute outcome reward:
\[
\begin{aligned}
x_i &= \mathcal{R}_{\mathrm{out}}(q,\mathcal{F}_i),\\
r_i^{\mathrm{out}}
&=
\frac{\mathrm{clip}(x_i,L,U)-L}{U-L}.
\end{aligned}
\]
\State Initialize $r_i^{\mathrm{sur}}\gets 0$
\If{$\mathcal{T}_i$ exists and $\mathcal{C}(q)\neq\emptyset$}
    \State Score $\mathcal{T}_i$ with $\mathcal{J}_{\mathrm{chk}}$ and obtain mean score $\bar{s}_i$
    \State Compute raw checklist reward:
    \[
    r_i^{\mathrm{chk}}
    =
    \left[
    \frac{\bar{s}_i-6}{4}
    \right]_+ .
    \]
    \State Update EMA ratio:
    \[
    \begin{aligned}
    \gamma_i
    &=
    \min\!\left(
    \frac{r_i^{\mathrm{chk}}}
         {r_i^{\mathrm{out}}+\epsilon},
    M
    \right),\\
    \alpha
    &\leftarrow
    \mu\alpha+(1-\mu)\gamma_i .
    \end{aligned}
    \]
    \State Compute residual surplus:
    \[
    r_i^{\mathrm{sur}}
    =
    \left[
    r_i^{\mathrm{chk}}
    -
    \alpha r_i^{\mathrm{out}}
    \right]_+ .
    \]
\EndIf
\State Compute final reward:
\[
\begin{aligned}
r_i
=&\,
(1-\lambda_{\mathrm{fmt}})r_i^{\mathrm{out}}
+\lambda_{\mathrm{fmt}}r_i^{\mathrm{fmt}} \\
&+
\lambda_{\mathrm{TCR}}r_i^{\mathrm{sur}} .
\end{aligned}
\]
\State \Return $\mathrm{clip}(r_i,0,1)$
\end{algorithmic}
\end{algorithm}

\section{Experimental Details}
\label{app:experimental_details}

\subsection{Training Data}
\label{app:training_data}

We use the BPO training data~\cite{cheng2024black} for policy optimization. 
The BPO training set is constructed from four preference-annotated datasets, covering both human-annotated and model-generated preference signals.

Specifically, OASST1~\cite{kopf2023openassistant} is a crowd-sourced instruction-following dataset with human-annotated response quality ratings. 
For each instruction, the response with the highest score is used as the preferred response, and the response with the lowest score is used as the rejected response. 
HH-RLHF~\cite{bai2022training} provides human preferences over helpfulness and harmlessness. 
Chatbot Arena Conversations~\cite{zheng2023judging} contains pairwise human preferences collected from the Chatbot Arena platform. 
BPO also includes the comparison subset of Alpaca-GPT4~\cite{peng2023instruction}, where preferences are generated by GPT-4~\cite{achiam2023gpt}. 
To ensure data quality, only samples in which GPT-4 responses outperform text-davinci-003 responses are retained.
For our method, we construct a sample-specific thinking checklist for each training question using its pairwise preference annotation. 
The resulting checklist is stored offline and used during reward computation. 
During RL training, all sampled responses for the same question share the same checklist, ensuring that checklist-based rewards are comparable within each rollout group.

\subsection{Training Configuration}
\label{app:training_configuration}

All RL experiments are implemented with the \texttt{verl} framework~\cite{sheng2025hybridflow} following the DAPO training recipe~\cite{yu2025dapo}. 
Unless otherwise specified, we use the same training configuration across backbone models. 
The main training hyperparameters are summarized in Table~\ref{tab:training_config}.

\begin{table}
\centering
\small
\setlength{\tabcolsep}{5pt}
\renewcommand{\arraystretch}{0.95}
\begin{tabular}{lc}
\toprule
Configuration & Value \\
\midrule
RL framework & \texttt{verl} \\
RL algorithm & DAPO \\
Advantage estimator & GRPO \\
Responses per prompt $G$ & 8 \\
Training prompt batch size & 128 \\
Generation prompt batch size & 128 \\
PPO mini-batch size & 32 \\
PPO micro-batch size per GPU & 16 \\
Max prompt length & 512 \\
Max response length & 4096 \\
Loss aggregation & token-mean \\
Learning rate & $1\times 10^{-6}$ \\
LR warmup steps & 10 \\
Weight decay & 0.1 \\
Gradient clipping & 1.0 \\
KL in reward & disabled \\
KL loss & disabled \\
KL coefficient & 0.0 \\
Entropy coefficient & 0.0 \\
Clip ratio low / high & $0.2 / 0.24$ \\
Dynamic sampling & disabled \\
Overlong buffer & disabled \\
Rollout backend & vLLM \\
Rollout temperature & 1.0 \\
Rollout top-$p$ & 1.0 \\
Rollout top-$k$ & -1 \\
Validation temperature & 0.6 \\
Validation top-$p$ & 0.95 \\
Number of nodes / GPUs & $1 / 2$ \\
\bottomrule
\end{tabular}
\caption{Main RL training hyperparameters.}
\label{tab:training_config}
\end{table}

We use Skywork-Reward-V2-Llama-3.1-8B \cite{liu2024skywork} as the outcome reward model and Qwen3-30B-A3B-Instruct-2507-FP8 \cite{qwen3technicalreport} as the checklist scoring model. 
The raw outcome reward is clipped to $[-25,70]$ and linearly normalized to $[0,1]$. 
For checklist scoring, the judge model assigns each checklist item one of five scores, $\{2,4,6,8,10\}$, where $6$ corresponds to the adequate level. 
The checklist reward is centered by $(s-6)/4$ and only the non-negative part is retained. 
For EMA-based residualization, we use a decay factor $\mu=0.99$, a numerical constant $\epsilon=10^{-6}$, and clip the checklist-to-outcome reward ratio by a maximum value of $5.0$. 
The format reward weight is set to $\lambda_{\mathrm{fmt}}=0.1$, and the TCR weight is set to $\lambda_{\mathrm{TCR}}=0.05$.

We use the same reward-related hyperparameters and normalization settings across all backbone models, without model-specific tuning. 
These values are selected based on the default DAPO training recipe and common RL post-training practice. 
This unified configuration allows us to evaluate whether TCR serves as a generally effective reward component, rather than a setting heavily tuned for a particular model or benchmark. 
In practical applications, these hyperparameters can be further adjusted according to the target model, task domain, and computational budget.
\subsection{Evaluation Benchmarks}
\label{app:evaluation_benchmarks}

We evaluate the trained models on four open-ended instruction-following benchmarks: Vicuna Eval, Dolly Eval, BPO-test Eval, and AlpacaEval~2.0.

\paragraph{Vicuna Eval.}
Vicuna Eval~\cite{chiang2023vicuna} contains 80 diverse questions covering eight categories. 
It is commonly used to evaluate open-ended instruction-following and conversational ability.

\paragraph{Dolly Eval.}
Dolly Eval~\cite{conover2023free} is a subset of 200 instances randomly sampled from the Dolly dataset. 
The original Dolly dataset is human-generated and covers eight categories of instruction-following tasks.

\paragraph{BPO-test Eval.}
BPO-test Eval is a 200-instance subset sampled from the test set of BPO~\cite{cheng2024black}. We use it to measure open-ended alignment performance. 

\paragraph{AlpacaEval 2.0.}
AlpacaEval~2.0~\cite{li2023alpacaeval} is an automatic evaluation benchmark for instruction-following models. 
We report the official \texttt{win\_rate}, \texttt{discrete\_win\_rate}, and \texttt{length\_controlled\_winrate}. 
The length-controlled win rate is used to mitigate the effect of response length in automatic pairwise evaluation.

\subsection{Evaluation Metrics}
\label{app:eval_metrics}

For AlpacaEval~2.0, we report three official metrics: \texttt{win\_rate}, \texttt{discrete\_win\_rate}, and \texttt{length\_controlled\_winrate}.
\texttt{win\_rate} measures the average preference probability that the evaluated model response is preferred over the baseline response.
\texttt{discrete\_win\_rate} converts each pairwise comparison into a discrete win/loss outcome and reports the proportion of wins.
\texttt{length\_controlled\_winrate} adjusts the win rate by controlling for response length, reducing the effect of verbosity on open-ended evaluation.
In the main text, we abbreviate these metrics as WR, DWR, and LCWR, respectively.

\subsection{Bootstrap Confidence Intervals}
\label{app:bootstrap_ci}

To further assess the reliability of the pairwise evaluation results, we conduct a bootstrap analysis for DAPO+TCR against DAPO. 
It is worth noting that the $\Delta$WR reported in Table~\ref{tab:pairwise_main} is computed as the average of the win-rate differences across Vicuna Eval, Dolly Eval, and BPO-test Eval. 
In contrast, the bootstrap confidence intervals reported here are computed over pooled evaluation instances from the three benchmarks. 
Specifically, for each backbone model, we merge all pairwise comparison instances from Vicuna Eval, Dolly Eval, and BPO-test Eval, and perform 10,000 bootstrap resampling runs over the pooled instances with replacement. 
For each resampled set, we compute the pooled $\Delta$WR, and then take the 2.5th and 97.5th percentiles as the 95\% confidence interval.

As shown in Table~\ref{tab:bootstrap_ci}, the confidence intervals are above zero for Qwen2.5-3B, Qwen2.5-7B, and Llama3.2-3B, indicating statistically reliable improvements on these models. 
For Llama3.1-8B and DeepSeek-LLM-7B, the intervals slightly overlap zero, while their point estimates in Table~\ref{tab:pairwise_main} remain positive. 
Overall, these results provide additional statistical support for the effectiveness of TCR.

\begin{table}[t]
\centering
\small
\begin{tabular}{lc}
\toprule
\textbf{Base LLM} & \textbf{95\% CI of pooled $\Delta$WR} \\
\midrule
Qwen2.5-3B        & [0.21, 16.04] \\
Qwen2.5-7B        & [1.88, 17.54] \\
Llama3.2-3B       & [0.62, 16.67] \\
Llama3.1-8B       & [-0.63, 15.00] \\
DeepSeek-LLM-7B   & [-1.46, 14.61] \\
\bottomrule
\end{tabular}
\vspace{-5pt}
\caption{Bootstrap 95\% confidence intervals of pooled $\Delta$WR for DAPO+TCR against DAPO. 
The intervals are computed with 10,000 bootstrap resampling runs over the pooled pairwise evaluation instances from Vicuna Eval, Dolly Eval, and BPO-test Eval. }
\vspace{-10pt}
\label{tab:bootstrap_ci}
\end{table}

\section{Checklist Analysis}
\label{app:checklist_analysis}

To better understand the properties of the generated thinking checklists, we analyze their length distribution, the types of thinking behaviors they capture, and representative checklist examples. 
This analysis focuses on the checklists used during training, where each checklist is constructed from pairwise preference data and serves as a sample-specific supervision target for the reasoning trace.

\subsection{Checklist Length Distribution}
\label{app:checklist_length_distribution}

We first analyze the length distribution of the generated thinking checklists. 
All 13,827 training samples have non-empty checklists, resulting in 70,482 checklist items in total. 
Each sample contains 5.10 checklist items on average, with a median of 5 and a standard deviation of 0.53. 
The checklist length ranges from 3 to 7 items, and each item contains 8.41 words on average. 
In addition, the number of unique item types is close to the total number of checklist items, with 70,359 unique item types and only 105 duplicated item types, suggesting that the generated checklists are highly sample-specific rather than dominated by repeated templates.

As shown in Table~\ref{tab:checklist_length_distribution}, most samples contain five checklist items, while a smaller portion contains four or six items. 
This suggests that the checklist construction process produces compact and stable supervision targets, rather than overly long rubrics. 
Meanwhile, the variation from three to seven items indicates that the checklist is not generated from a rigid fixed-length template, but adapts moderately to the complexity of each sample.

\begin{table}[t]
\centering
\small
\setlength{\tabcolsep}{6pt}
\renewcommand{\arraystretch}{0.98}
\begin{tabular}{ccc}
\toprule
\# Checklist Items & Count & Percentage \\
\midrule
3 & 8 & 0.06\% \\
4 & 1,140 & 8.24\% \\
5 & 10,332 & 74.72\% \\
6 & 2,191 & 15.85\% \\
7 & 156 & 1.13\% \\
\bottomrule
\end{tabular}
\caption{Distribution of checklist lengths.}
\label{tab:checklist_length_distribution}
\end{table}

\begin{table}[t]
\centering
\small
\setlength{\tabcolsep}{3.5pt}
\renewcommand{\arraystretch}{0.95}
\begin{tabular}{lcc}
\toprule
Thinking Behavior & Count & Percentage \\
\midrule
Context Grounding & 927 & 18.54\% \\
User Adaptation & 862 & 17.24\% \\
Completeness Planning & 796 & 15.92\% \\
Logical Planning & 519 & 10.38\% \\
Answer Organization & 421 & 8.42\% \\
Trade-off Awareness & 394 & 7.88\% \\
Evidence Checking & 364 & 7.28\% \\
Intent Decomposition & 285 & 5.70\% \\
Risk/Safety Awareness & 225 & 4.50\% \\
Constraint Tracking & 207 & 4.14\% \\
\bottomrule
\end{tabular}

\caption{Distribution of thinking behavior categories captured by generated checklist items, computed on a random subset of 5,000 checklist items.}
\label{tab:checklist_category_distribution}
\end{table}

\begin{table*}[!t]
\centering
\small
\setlength{\tabcolsep}{4pt}
\renewcommand{\arraystretch}{1.05}
\begin{tabular}{p{0.25\linewidth}p{0.68\linewidth}}
\toprule
Thinking Behavior & Definition \\
\midrule
Intent Decomposition 
& Identifying, inferring, or decomposing the user's goal, intent, or task objective before generating the final answer. \\
Constraint Tracking 
& Tracking explicit requirements, formats, conditions, or constraints specified in the user request during reasoning. \\
Context Grounding 
& Grounding the reasoning process in the specific context, scenario, or background of the input question. \\
Evidence and Fact Checking 
& Checking factual claims, assumptions, evidence, and logical consistency before committing to the final response. \\
Logical Planning 
& Forming a coherent reasoning path, causal analysis, step-by-step plan, or analytical structure. \\
Completeness Planning 
& Identifying the key aspects that need to be covered and avoiding important omissions during planning. \\
Trade-off Awareness 
& Considering alternatives, limitations, pros and cons, or nuanced trade-offs when the task involves multiple possible directions. \\
Risk and Safety Awareness 
& Anticipating potential risks, sensitive issues, unsafe advice, bias, or inappropriate conclusions during reasoning. \\
Answer Organization Planning 
& Planning how the final answer should be structured, sequenced, or presented based on the reasoning outcome. \\
User Adaptation 
& Adapting the reasoning strategy to the user's needs, background, tone, expected level of detail, or practical use case. \\
\bottomrule
\end{tabular}
\caption{Definitions of thinking behavior categories used for checklist analysis. Each category describes a process-level behavior expected from the reasoning trace, rather than a surface-level rubric for the final answer.}
\label{tab:thinking_behavior_definitions}
\end{table*}
\subsection{Checklist Category Distribution}
\label{app:checklist_category_distribution}

\begin{table*}[!t]
\centering
\small
\setlength{\tabcolsep}{0pt}
\renewcommand{\arraystretch}{1.08}
\begin{tabular}{p{0.96\linewidth}}
\toprule
\textbf{Representative Generated Checklist Items by Thinking Behavior} \\
\midrule
\textbf{Intent Decomposition.}
User intent interpretation (clarify underlying needs without violating confidentiality norms).
Clarification of user intent (probe underlying needs or nuances in the question). \\[3pt]

\textbf{Constraint Tracking.}
Relevance to query scope (focus on exact word constraints: ``five letters'', ``al-'', ``-e'').
Environment compatibility (space needs and adaptability to specific living conditions). \\[3pt]

\textbf{Context Grounding.}
Contextual background (clarifies origins, goals, and intentions of the topic).
Historical context (timelines of adoption, obsolescence, and replacement). \\[3pt]

\textbf{Evidence and Fact Checking.}
Assumption validation (addressing implicit assumptions within the user's question).
Evidence alignment (considering scientific evidence and how it supports the argument). \\[3pt]

\textbf{Logical Planning.}
Alternative root causes (explore diverse potential sources beyond the immediate error context).
Logical connections (relationships between causes, consequences, and underlying principles). \\[3pt]

\textbf{Completeness Planning.}
Range of potential causes (breadth of contributing factors considered).
Consideration of exceptions (addresses edge cases or deviations from the norm). \\[3pt]

\textbf{Trade-off Awareness.}
Plausibility of multiple hypotheses (weigh competing explanations across different contexts and evidence).
Survival trade-offs (weighing benefits like protection against potential disadvantages or risks). \\[3pt]

\textbf{Risk and Safety Awareness.}
Ethical implications of knowledge sharing (evaluate risks of harmful misuse of information).
Risk management focus (consider identification and mitigation of potential hazards). \\[3pt]

\textbf{Answer Organization Planning.}
Transparency of reasoning (consider whether steps are explicitly outlined for comprehension).
Hierarchical clarity (organizes ideas, distinguishes broader concepts from specific examples). \\[3pt]

\textbf{User Adaptation.}
Problem anticipation (proactive thinking about potential user needs or clarifications).
User comprehension focus (attentiveness to anticipated audience knowledge and needs). \\
\bottomrule
\end{tabular}
\caption{Representative generated checklist items for each thinking behavior category.}
\label{tab:checklist_examples}
\end{table*}

We further analyze what types of thinking behaviors are captured by the generated checklists. 
Since the checklist is designed to supervise reasoning traces rather than final responses, we categorize checklist items according to the reasoning behavior they require from the model's thinking process.

We classify a random subset of 5,000 checklist items using this taxonomy. 
As shown in Table~\ref{tab:checklist_category_distribution}, the generated checklists cover a diverse set of thinking behaviors. 
The most frequent categories are context grounding, user adaptation, and completeness planning, accounting for 18.54\%, 17.24\%, and 15.92\% of the analyzed items, respectively. 
This suggests that the checklists often encourage the model to anchor its reasoning in the specific input, adapt its reasoning strategy to user needs, and plan the key aspects to cover before answering. 
Other categories, such as logical planning, answer organization planning, trade-off awareness, and evidence checking, further indicate that the checklists supervise structured reasoning behaviors rather than generic response-level qualities.
Table~\ref{tab:thinking_behavior_definitions} defines the thinking-behavior taxonomy used in this analysis.

\subsection{Judge Consistency}
\label{app:judge_consistency}

To examine whether checklist-based scoring is overly dependent on a single judge model, we randomly sample 500 responses and ask different LLM judges to score the same reasoning traces with the same generated checklists. 
The scores are generally consistent across judges. 
This suggests that checklist-based scoring is a relatively simple and constrained judging task, because the checklist already specifies what thinking behaviors should be checked. 
Thus, the judge only needs to assess whether the reasoning trace satisfies these explicit criteria, rather than make an unconstrained preference judgment.
\begin{table*}[h]
\centering
\small
\setlength{\tabcolsep}{0pt}
\begin{tabular}{p{0.96\linewidth}}
\toprule
\textbf{Checklist Construction Prompt} \\
\midrule
\begin{minipage}{0.96\linewidth}
\begin{Verbatim}[breaklines=true, fontsize=\scriptsize]
You are an experienced data annotator specializing in creating high-quality training data for reward models. 
Your task is to analyze the following input:

[User Question]
{user_question}
[Better Answer]
{good_res}
[Weaker Answer]
{bad_res}

Task:
- From the two answers, infer the **thinking dimensions** the better answer attends to that the weaker answer overlooks or underdevelops.
- Convert these into a **thinking checklist** that guides what to think about (cognitive focuses), not what to do (procedures).
- Each item should be a concise **noun phrase or evaluation criterion** that names the mental consideration, optionally followed by a brief clarifier (<= 12 words).

Output Format (strict):
Thinking Checklist:
 Key point 1: <thinking dimension / criterion (optional brief clarifier)>
 Key point 2: <thinking dimension / criterion (optional brief clarifier)>
 ...

Additional Requirements:
- Do not mention the existence of two answers or compare them explicitly.
- Focus on **what to examine/consider/evaluate**, not how to write/format/structure steps.
- Avoid procedural verbs like “write”, “explain”, “list”, “provide”, “format”, “walk through”.
- Keep items generalizable and domain-relevant; 2–7 items total; each on one line.
\end{Verbatim}
\end{minipage}
\\
\bottomrule
\end{tabular}
\caption{Prompt used to construct thinking checklists from pairwise preference data.}
\label{tab:checklist_construction_prompt}
\end{table*}

\begin{table*}[!t]
\centering
\small
\setlength{\tabcolsep}{0pt}
\begin{tabular}{p{0.96\linewidth}}
\toprule
\textbf{Pairwise Evaluation Prompt} \\
\midrule
\begin{minipage}{0.96\linewidth}
\begin{Verbatim}[breaklines=true, fontsize=\scriptsize]
System prompt:

Please act as an impartial judge and evaluate the quality of the responses provided by two AI assistants to the user question displayed below. You should choose the assistant that follows the user's instructions and answers the user's question better. Your evaluation should consider factors such as the helpfulness, relevance, accuracy, depth, creativity, and level of detail of their responses. Begin your evaluation by comparing the two responses and provide a short explanation. Avoid any position biases and ensure that the order in which the responses were presented does not influence your decision. Do not allow the length of the responses to influence your evaluation. Do not favor certain names of the assistants. Be as objective as possible. After providing your explanation, output your final verdict by strictly following this format: "[[A]]" if assistant A is better, "[[B]]" if assistant B is better, and "[[C]]" for a tie.

Prompt template:

[User Question]
{question}

[The Start of Assistant A's Answer]
{answer_a}
[The End of Assistant A's Answer]

[The Start of Assistant B's Answer]
{answer_b}
[The End of Assistant B's Answer]

Output format:
[[A]]
\end{Verbatim}
\end{minipage}
\\
\bottomrule
\end{tabular}
\caption{Prompt used for pairwise response evaluation. The order of responses is randomly shuffled during evaluation to reduce position bias.}
\label{tab:pairwise_evaluation_prompt}
\end{table*}

\subsection{Examples of Generated Checklist Items}
\label{app:checklist_examples}

Table~\ref{tab:checklist_examples} presents representative generated checklist items for the ten thinking behavior categories. 
All examples are selected from the analyzed checklist items without rewriting. 
We choose items that explicitly describe process-level considerations for the reasoning trace, such as clarifying intent, tracking constraints, validating assumptions, considering alternatives, and anticipating risks.

\section{Prompts}
\label{app:prompts}

This section lists the prompts and checklist used in our experiments.
Specifically, we provide four prompts for checklist construction, checklist-based reasoning evaluation, generic trajectory reward evaluation, and pairwise response evaluation, together with the unified global checklist used by the Global Checklist Reward baseline.

\begin{table*}[!t]
\centering
\small
\setlength{\tabcolsep}{0pt}
\begin{tabular}{p{0.96\linewidth}}
\toprule
\textbf{Checklist Scoring Prompt} \\
\midrule
\begin{minipage}{0.96\linewidth}
\begin{Verbatim}[breaklines=true, fontsize=\scriptsize]
You are a strict and conservative evaluator. Your goal is to provide stable, low-variance,
and evidence-based judgments.

IMPORTANT: Scoring must follow a TWO-STEP process.

STEP 1 — Choose exactly ONE rating level from the following list:
- "very poor"
- "poor"
- "adequate"
- "good"
- "excellent"

STEP 2 — Map the chosen rating level to a numeric score using this fixed mapping:
- very poor → 2
- poor → 4
- adequate → 6
- good → 8
- excellent → 10

For each checklist item:
1) First identify concrete flaws, gaps, or weaknesses.
2) Briefly explain how well the {subject_type.lower()} satisfies the criterion (1–2 sentences).
3) Choose ONE rating level.
4) Output the corresponding numeric score.

After all items:
- Provide an overall summary (3–4 sentences).
- Compute overall_score as the arithmetic mean of all numeric scores, rounded to one decimal.

Output MUST be valid JSON only, exactly following this schema:
{
  "items": [
    {
      "criterion": "<text>",
      "explanation": "<1-2 sentences>",
      "rating_level": "<very poor | poor | adequate | good | excellent>",
      "score": <2 | 4 | 6 | 8 | 10>
    }
  ],
  "overall_summary": "<3-4 sentences>",
  "overall_score": <number with one decimal>
}

If the checklist is empty, return:
- "items": []
- "overall_score": 0.0
with an appropriate overall_summary.

--- {subject_type} ---
{subject_text}

--- Checklist ---
{checklist_display}

Return valid JSON only. No extra text.
\end{Verbatim}
\end{minipage}
\\
\bottomrule
\end{tabular}
\caption{Prompt used to score a generated reasoning trace against the thinking checklist.}
\label{tab:checklist_scoring_prompt}
\end{table*}

\begin{table*}[!t]
\centering
\small
\setlength{\tabcolsep}{0pt}
\begin{tabular}{p{0.96\linewidth}}
\toprule
\textbf{Generic Trajectory Reward Prompt} \\
\midrule
\begin{minipage}{0.96\linewidth}
\begin{Verbatim}[breaklines=true, fontsize=\scriptsize]
You are a strict and conservative evaluator of reasoning quality.

Given a user question and a model-generated reasoning trace, evaluate the overall quality of the reasoning process.

IMPORTANT:
- Do NOT evaluate the final answer itself.
- Do NOT reward verbosity by itself.
- Only evaluate whether the reasoning trace is useful, coherent, relevant, and likely to support a high-quality answer.

Evaluate the reasoning trace according to the following aspects:
1) Whether it correctly understands the user's intent.
2) Whether it identifies relevant requirements, constraints, and context.
3) Whether it develops a coherent and useful reasoning process.
4) Whether it avoids unsupported assumptions, hallucinations, and irrelevant reasoning.
5) Whether it is logically consistent and helpful for producing a good final answer.

Choose exactly ONE rating level:
- "very poor"
- "poor"
- "adequate"
- "good"
- "excellent"

Map the rating level to a numeric score:
- very poor -> 2
- poor -> 4
- adequate -> 6
- good -> 8
- excellent -> 10

Output MUST be valid JSON only, exactly following this schema:
{
  "explanation": "<brief explanation in 2-3 sentences>",
  "rating_level": "<very poor | poor | adequate | good | excellent>",
  "score": <2 | 4 | 6 | 8 | 10>
}

--- User Question ---
{question}

--- Reasoning Trace ---
{think}

Return valid JSON only. No extra text.
\end{Verbatim}
\end{minipage}
\\
\bottomrule
\end{tabular}
\caption{
Prompt used by the Generic Traj. Reward baseline to directly score the whole reasoning trace without checklist guidance.
}
\label{tab:generic_traj_reward_prompt}
\end{table*}

\begin{table*}[!t]
\centering
\small
\setlength{\tabcolsep}{0pt}
\begin{tabular}{p{0.96\linewidth}}
\toprule
\textbf{Global Thinking Checklist} \\
\midrule
\begin{minipage}{0.96\linewidth}
\begin{Verbatim}[breaklines=true, fontsize=\scriptsize]
1. Correctly understand the user's intent and the main task.
2. Identify the key requirements, constraints, and conditions in the question.
3. Develop a coherent and useful reasoning plan before answering.
4. Use relevant information from the question and avoid irrelevant reasoning.
5. Avoid unsupported assumptions, hallucinated facts, or overgeneralization.
6. Consider important trade-offs, edge cases, or contextual factors when needed.
7. Ensure the reasoning process is logically consistent and leads toward a helpful answer.
\end{Verbatim}
\end{minipage}
\\
\bottomrule
\end{tabular}
\caption{
Global thinking checklist used by the Global Checklist Reward baseline.
}
\label{tab:global_thinking_checklist}
\end{table*}

\section{Case Study}
\label{app:case_study}

We provide case studies comparing models trained with and without TCR. 
The detailed examples are presented in Tables~\ref{tab:case_study_1}, \ref{tab:case_study_2}, and~\ref{tab:case_study_3}.

\begin{table*}[!t]
\centering
\small
\setlength{\tabcolsep}{0pt}
\begin{tabular}{p{0.96\linewidth}}
\toprule
\textbf{Case Study 1: Challenges in the Education Sector} \\
\midrule
\begin{minipage}{0.96\linewidth}
\textbf{Question.}
What are the major challenges faced by the education sector today?

\vspace{2pt}
\textbf{\textcolor{red!70!black}{Reasoning w/o TCR.}}
\begin{Verbatim}[breaklines=true, fontsize=\scriptsize]
To address the major challenges faced by the education sector today, I need to consider several key areas that have been impacted by various factors such as technological advancements, socio-economic changes, and global events. These areas include access to education, quality of education, teacher training, curriculum relevance, and the integration of technology.
\end{Verbatim}

\vspace{2pt}
\textbf{\textcolor{red!70!black}{Answer w/o TCR.}}
\begin{Verbatim}[breaklines=true, fontsize=\scriptsize]
The education sector today faces several significant challenges:

1. **Access to Education**: Despite progress, many regions still struggle with providing equitable access to education. [...]
2. **Quality of Education**: Ensuring that all students receive a high-quality education that prepares them for the future is a major challenge. [...]
3. **Teacher Training and Support**: Teachers need continuous professional development [...]
4. **Curriculum Relevance**: The curriculum needs to evolve to reflect the changing needs of society and the workforce. [...]
5. **Integration of Technology**: While technology has the potential to enhance learning, its integration into education faces challenges [...]
6. **Global Events and Societal Changes**: Events such as pandemics, economic downturns, and political instability can disrupt education systems [...]
\end{Verbatim}

\vspace{2pt}
\textbf{\textcolor{blue!70!black}{Reasoning w/ TCR.}}
\begin{Verbatim}[breaklines=true, fontsize=\scriptsize]
To address the major challenges faced by the education sector today, I need to consider several key areas that have been impacted by various factors such as technological advancements, socio-economic changes, and global events. Here are the steps I'll take:

1. Identify broad categories of challenges.
2. Provide specific examples under each category.
3. Consider the impact of these challenges on different stakeholders (students, educators, administrators).
4. Ensure the answer is comprehensive and covers the most significant issues.
\end{Verbatim}

\vspace{2pt}
\textbf{\textcolor{blue!70!black}{Answer w/ TCR.}}
\begin{Verbatim}[breaklines=true, fontsize=\scriptsize]
The education sector today faces several significant challenges that impact its effectiveness and accessibility. Here are some of the major challenges:

1. **Technological Integration and Accessibility**:
   - **Digital Divide**: Not all students have equal access to technology and the internet [...]
   - **Digital Literacy**: Educators and students need to develop skills to effectively use digital tools and platforms.

2. **Curriculum Relevance and Adaptability**:
   - **Changing Job Market**: The rapid pace of technological change requires curricula to be updated frequently [...]
   - **Global Competitiveness**: Educational systems must adapt to compete globally [...]

3. **Educational Inequality**:
   - **Resource Allocation**: Schools in affluent areas often have better resources [...]
   - **Social and Economic Barriers**: Students from lower socio-economic backgrounds may face additional challenges [...]

4. **Teacher Quality and Professional Development**:
   - **Retention and Recruitment** [...]
   - **Professional Growth** [...]

5. **Assessment and Evaluation**:
   - **Standardized Testing** [...]
   - **Holistic Assessment** [...]

6. **Mental Health and Well-being**:
   - **Stress and Anxiety** [...]
   - **Support Services** [...]

7. **Global Events and Disruptions**:
   - **Pandemics and Natural Disasters** [...]
   - **Political and Economic Instability** [...]
\end{Verbatim}

\vspace{2pt}
\textbf{Observation.}
The model trained w/ TCR first plans how to organize the analysis, including category construction, examples, stakeholder impact, and coverage. This planning is reflected in the final answer, which covers more concrete and contemporary challenges such as the digital divide, digital literacy, educational inequality, assessment, mental health, and disruption risks.
\end{minipage}
\\
\bottomrule
\end{tabular}
\caption{Case study on education-sector challenges. The model trained w/ TCR produces more planned reasoning and a more comprehensive final answer.}
\label{tab:case_study_1}
\end{table*}

\begin{table*}[!t]
\centering
\small
\setlength{\tabcolsep}{0pt}
\begin{tabular}{p{0.96\linewidth}}
\toprule
\textbf{Case Study 2: Automation, Job Creation, and Technological Progress} \\
\midrule
\begin{minipage}{0.96\linewidth}
\textbf{Question.}
In a world where automation is becoming increasingly prevalent, is it more important to prioritize job creation or technological progress?

\vspace{2pt}
\textbf{\textcolor{red!70!black}{Reasoning w/o TCR.}}
\begin{Verbatim}[breaklines=true, fontsize=\scriptsize]
To determine whether job creation or technological progress should be prioritized in a world where automation is becoming increasingly prevalent, I need to consider the benefits and drawbacks of each approach.

1. Technological progress:
   - Innovation drives economic growth and can lead to new industries and job opportunities.
   - Automation can increase efficiency and productivity [...]
   - However, rapid technological change can also lead to job displacement and income inequality.

2. Job creation:
   - Ensuring that people have employment can improve quality of life and reduce poverty.
   - Job creation can provide a sense of purpose and fulfillment [...]
   - However, simply creating jobs does not guarantee that they are sustainable or well-paying.

The challenge is to balance these factors.
\end{Verbatim}

\vspace{2pt}
\textbf{\textcolor{red!70!black}{Answer w/o TCR.}}
\begin{Verbatim}[breaklines=true, fontsize=\scriptsize]
In a world where automation is becoming increasingly prevalent, it is important to balance the priorities of job creation and technological progress. While technological progress is essential for long-term economic growth and can lead to the development of new industries, it is also crucial to manage the transition in a way that mitigates job displacement and income inequality. Simultaneously, job creation is important to ensure that people have meaningful and well-compensated work. The key is to invest in education and retraining programs [...]
\end{Verbatim}

\vspace{2pt}
\textbf{\textcolor{blue!70!black}{Reasoning w/ TCR.}}
\begin{Verbatim}[breaklines=true, fontsize=\scriptsize]
The question of whether to prioritize job creation or technological progress in a world dominated by automation is complex and multifaceted. Let's break it down:

1. Economic Growth and Innovation: Technological progress often drives economic growth and innovation. New technologies can create entirely new industries and job sectors [...]

2. Job Displacement: Automation can also lead to job displacement in certain sectors. Jobs that are repetitive or can be easily automated might become obsolete [...]

3. Skill Mismatch: As technology advances, there is often a mismatch between the skills that are in demand and the skills that the workforce possesses. This can exacerbate inequality and economic disparity.

4. Long-term Benefits [...]
5. Policy and Education [...]
6. Sustainability and Ethical Considerations [...]

Given these considerations, it seems that both job creation and technological progress are important, but they are not mutually exclusive.
\end{Verbatim}

\vspace{2pt}
\textbf{\textcolor{blue!70!black}{Answer w/ TCR.}}
\begin{Verbatim}[breaklines=true, fontsize=\scriptsize]
In a world where automation is becoming increasingly prevalent, it is crucial to strike a balance between prioritizing job creation and technological progress. Technological progress is essential for driving economic growth, innovation, and addressing global challenges like climate change. However, it is equally important to ensure that this progress does not lead to significant job displacement and skill mismatches. By implementing effective policies and education systems that support workforce retraining and upskilling, we can harness the benefits of technological advancement while creating new job opportunities. The key is to foster an environment where technological progress and job creation are mutually reinforcing, rather than competing, goals.
\end{Verbatim}

\vspace{2pt}
\textbf{Observation.}
The model trained w/ TCR moves beyond a simple two-sided comparison and explicitly reasons about skill mismatch, policy and education, and sustainability and ethical issues. The final answer also reflects this improved reasoning by reframing job creation and technological progress as mutually reinforcing goals rather than competing priorities.
\end{minipage}
\\
\bottomrule
\end{tabular}
\caption{Case study on trade-off reasoning under automation. The model trained w/ TCR better captures the interaction between technological progress, job displacement, and policy responses.}
\label{tab:case_study_2}
\end{table*}

\begin{table*}[!t]
\centering
\small
\setlength{\tabcolsep}{0pt}
\begin{tabular}{p{\linewidth}}
\toprule
\textbf{Case Study 3: Podcast Script on Streaming Platforms} \\
\midrule
\begin{minipage}{\linewidth}
\textbf{Question.}
Structure a podcast script for an episode discussing the influence of streaming platforms on the music industry.

\vspace{2pt}
\textbf{\textcolor{red!70!black}{Reasoning w/o TCR.}}
\begin{Verbatim}[breaklines=true, fontsize=\scriptsize]
To structure a podcast episode on the influence of streaming platforms on the music industry, I need to consider the key points that would be interesting and informative for listeners. The episode should cover the rise of streaming, its impact on album sales, artist discovery, and revenue models. It's also important to include perspectives from both industry insiders and artists themselves. The script should flow logically, starting with an introduction, moving through the main points, and ending with a conclusion or call to action.
\end{Verbatim}

\vspace{2pt}
\textbf{\textcolor{red!70!black}{Answer w/o TCR.}}
\begin{Verbatim}[breaklines=true, fontsize=\scriptsize]
[intro music plays]
Host: [welcoming tone] Welcome to "The Soundtrack of Our Lives," your go-to podcast for all things music and technology. [...]

[Transition music]
Host: First, let's talk about the rise of streaming platforms. Services like Spotify, Apple Music, and Tidal have grown exponentially since their inception. [...]

[Transition music]
Host: One of the most significant impacts of streaming has been on album sales. [...]

[Transition music]
Host: Let's hear from a real artist about this change. [Name of artist], a [genre] artist, has been in the industry for over a decade. [...]

[Transition music]
Host: Streaming platforms have made it easier than ever to discover new artists. [...]

[Transition music]
Host: As we wrap up today's episode, it's clear that streaming platforms have had a profound impact on the music industry. [...]
\end{Verbatim}

\vspace{2pt}
\textbf{\textcolor{blue!70!black}{Reasoning w/ TCR.}}
\begin{Verbatim}[breaklines=true, fontsize=\scriptsize]
To structure a podcast episode on the influence of streaming platforms on the music industry, I need to consider the key points that would be interesting and informative for listeners. The episode should cover the rise of streaming, its impact on album sales, artist discovery, and revenue models. It's also important to discuss the challenges and opportunities that streaming presents to both established and up-and-coming artists. I should include examples of successful artists who have thrived on streaming platforms and perhaps some case studies of artists who have struggled. Additionally, I should touch on the broader implications for the industry, such as changes in how music is produced and consumed. The episode should be engaging and accessible, so I'll aim for a conversational tone and include anecdotes or interviews with industry experts.
\end{Verbatim}

\vspace{2pt}
\textbf{\textcolor{blue!70!black}{Answer w/ TCR.}}
\begin{Verbatim}[breaklines=true, fontsize=\scriptsize]
[Podcast Intro Music]
Host: [Warm, inviting tone] Welcome to "The Soundtrack of Our Lives," where we explore the music industry and its impact on our daily lives. [...]

[Transition Music]
Host: Let's start by looking at how streaming platforms like Spotify, Apple Music, and Tidal have reshaped the music industry. Before streaming, the primary way to listen to music was through physical media like CDs and vinyl, or through radio. But with the advent of streaming, the way we consume music has shifted dramatically. [...]

[Segment Transition Music]
Host: One of the most significant impacts of streaming has been on album sales. With the rise of streaming, the concept of the "album" as a cohesive unit has become less relevant. Listeners can now access individual songs and playlists, which has led to a shift in how artists release music. [...]

[Transition Music]
Host: Speaking of standing out, let's talk about how streaming has changed the way artists are discovered. [...]

[Segment Transition Music]
Host: However, the shift to streaming has also brought challenges. For one, the revenue model for artists is different. While streaming platforms pay artists per stream, the payouts are often small. This has led to debates about fair compensation and the value of individual streams. [...]

[Transition Music]
Host: To get a deeper understanding of these issues, I spoke with [Name of Industry Expert], a music industry analyst and former record label executive. [...]

Host: That's a great point. It's clear that while streaming has opened up new opportunities, it's also created new challenges. [...]
\end{Verbatim}

\vspace{2pt}
\textbf{Observation.}
The model trained w/ TCR plans the script around a richer set of industry-level considerations, including opportunities and challenges for different artists, case examples, expert input, and changes in music production and consumption. These considerations appear in the final script through more concrete segments on consumption shifts, album unbundling, revenue challenges, fair compensation, and industry adaptation.
\end{minipage}
\\
\bottomrule
\end{tabular}
\caption{Case study on podcast-script generation. The model trained w/ TCR produces more comprehensive reasoning and a richer final script structure.}
\label{tab:case_study_3}
\end{table*}

\end{document}